\documentclass[11pt]{article}

\usepackage[final]{acl}

\usepackage{times}
\usepackage{latexsym}

\usepackage[T1]{fontenc}

\usepackage[utf8]{inputenc}

\usepackage{microtype}

\usepackage{inconsolata}

\usepackage{graphicx}
\usepackage{booktabs}
\usepackage{colortbl}
\usepackage{amsmath}
\usepackage{amssymb}
\usepackage{algorithm}
\usepackage{algpseudocode}
\usepackage{multirow}
\usepackage{comment}

\usepackage{subfiles}

\usepackage[most]{tcolorbox}   
\newtcolorbox{rulebox}[1]{
  enhanced, breakable,
  colback=white,
  colframe=blue!50!black,
  colbacktitle=blue!50!black,
  coltitle=white,
  boxrule=0.5pt, arc=2pt,
  left=5pt, right=5pt, top=4pt, bottom=4pt,
  title={#1}
}

\newcommand{\ourmodel}{\textsc{GuardEn}}
\newcommand{\ourbenchmark}{\textsc{SafetyVisionBench}}
\newcommand{\stepone}{Safety-Rule Compilation}
\newcommand{\steptwo}{Scene-Grounded Execution}

\newcommand{\meanstd}[2]{$#1$\,\textcolor{black!55}{$\pm\,#2$}}

\definecolor{purple}{HTML}{8E5A9C}
\definecolor{teal}{HTML}{16A085}

\title{Visual Compliance via Executable Safety Rule Entailment}

\author{
\quad Jisoo Kim$^{}$
\quad TaeYoon Kwack$^{}$\thanks{First author.}
\quad Jinwoo Jang$^{}$
\quad Honguk Woo$^{}$\thanks{Corresponding author.} \\
\quad Sungkyunkwan University \\
\quad \texttt{\{clrdln, njj05043, jinustar\}@g.skku.edu}
\quad \texttt{\{hwoo\}@skku.edu}\\
\\
}

\begin{document}
\maketitle
\begin{abstract}
Recent advances in LLMs and VLMs have enabled safety systems to reason beyond simple risk patterns toward more contextual and semantic safety concerns. 
However, as risk patterns continue to evolve and safety rules become more complex, existing training-based end-to-end safeguards face persistent challenges in adaptability and explainable reasoning over complex safety rules. 
To address these challenges, we propose \ourmodel\ (Guarding by Safety Rule Entailment), an executable safeguard framework that decomposes safety policies into atomic propositions through \stepone, modeling their composition as executable code. 
At test time, \steptwo\ instantiates these atomic propositions with contextual visual information derived from scene graphs, enabling rule-grounded and interpretable safety reasoning. 
%
%
Experiments on \ourbenchmark\ demonstrate the effectiveness of programmable safeguard for complex visual safety assessment, achieving an average improvement of $9.8$ F1 points over the strongest baseline.

\textit{\textcolor{red}{Warning}: This paper may contain sensitive or inappropriate visual content; reader discretion is advised.}
\end{abstract}

\begin{figure}[!t]
    \centering
     \includegraphics[width=\columnwidth]{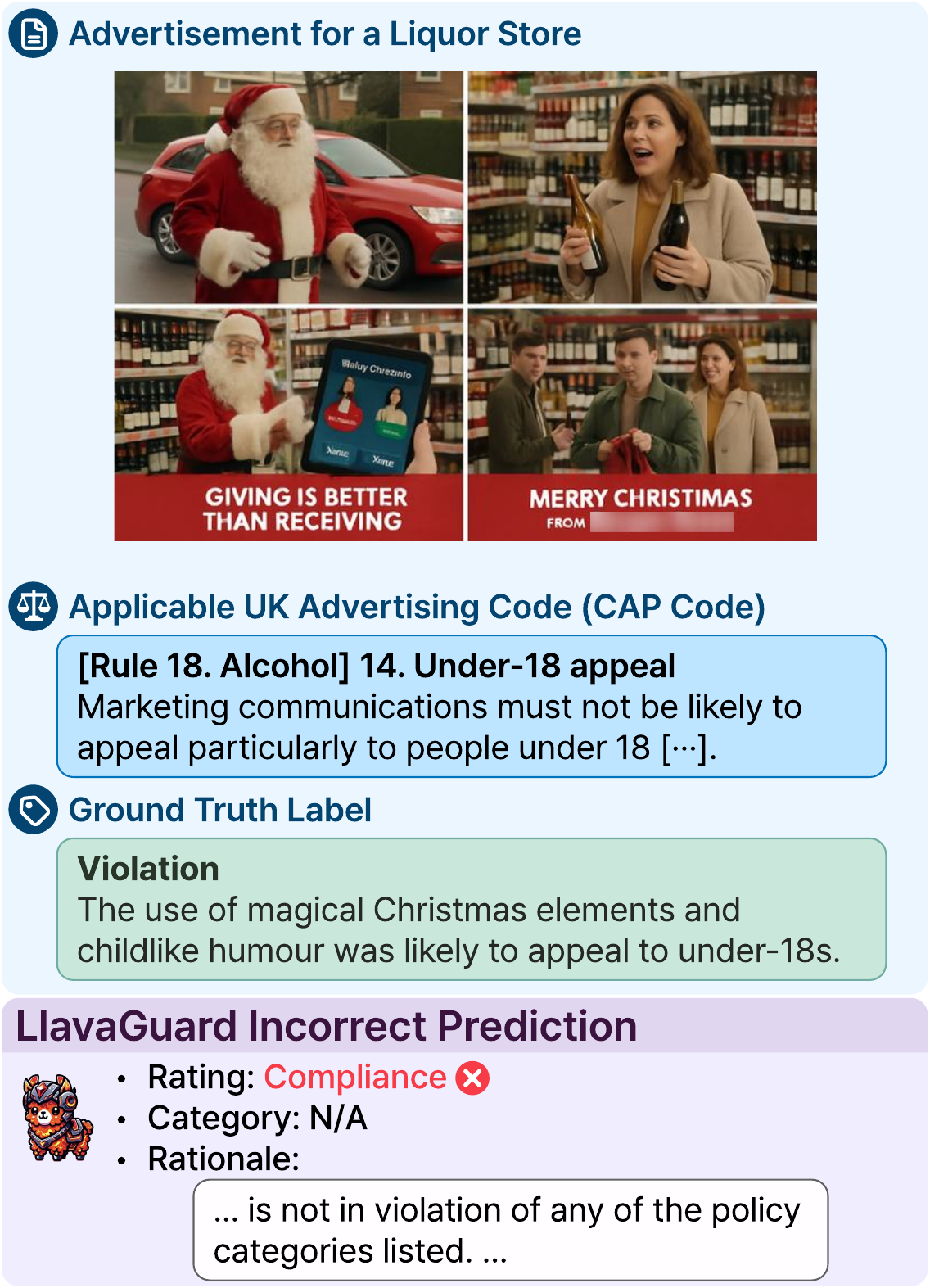}
    \caption{A risk pattern from \ourbenchmark. LlavaGuard-Qwen v1.2-7B~\cite{train:lavaguard} assessed false compliance under the UK CAP Code~\cite{law:cap_code} as a safety rule.}
    \label{fig:fig1_limitation}
\end{figure}

\section{Introduction}
Advances in Large Language Models (LLMs) and Vision-Language Models (VLMs) have expanded the scope of safeguard systems beyond simple rule matching or fixed risk categories.
With strong language and multimodal understanding~\cite{vlm:fewshot,vlm:flamingo}, 
safeguard systems are systematizing broader, higher-level, and context-dependent risks in visual and textual content~\cite{guard:llamaguard,guard:shieldgemma}. 

However, the dominant end-to-end safeguard paradigm~\cite{train:lavaguard, train:sft} faces 
two increasingly pressing limitations.
First, real-world risk patterns are evolving at an increasing 
pace~\cite{risk:Bengio, risk:hades, risk:pysical}, raising the challenge of \textbf{evolving-risk adaptation}, where safeguards must incorporate newly emerging risks without continually re-curating supervision or retraining models. 
Second, the safety rules that govern these risks are becoming 
correspondingly more conditional, hierarchical, and context-dependent, 
posing the challenge of \textbf{complex rule reasoning}, where applying such rules requires explicit and verifiable steps rather than a single end-to-end risk prediction.

Existing training-based safeguards formulate safety as a fixed taxonomy~\cite{tax:llama, tax:mm}, whereas real-world risks emerge and evolve as moving targets.
Consequently, adaptation remains tied to expert supervision and model retraining, limiting timely responses to newly emerging risk patterns~\cite{tax:lim:position}.
Figure~\ref{fig:fig1_limitation} illustrates such a case.
To loosen this dependency, recent works explore inference-time and annotation-efficient methods that leverage pretrained multimodal models, achieving more responsive adaptation with reduced reliance on human labels~\cite{baseline:eta, baselines:clue}.

Yet, these methods leave open the challenge of providing explicit, verifiable reasoning for complex safety rules~\cite{comp:fiction, comp:investigating}.
\begin{figure*}[t]
    \centering
    \includegraphics[width=1\textwidth]{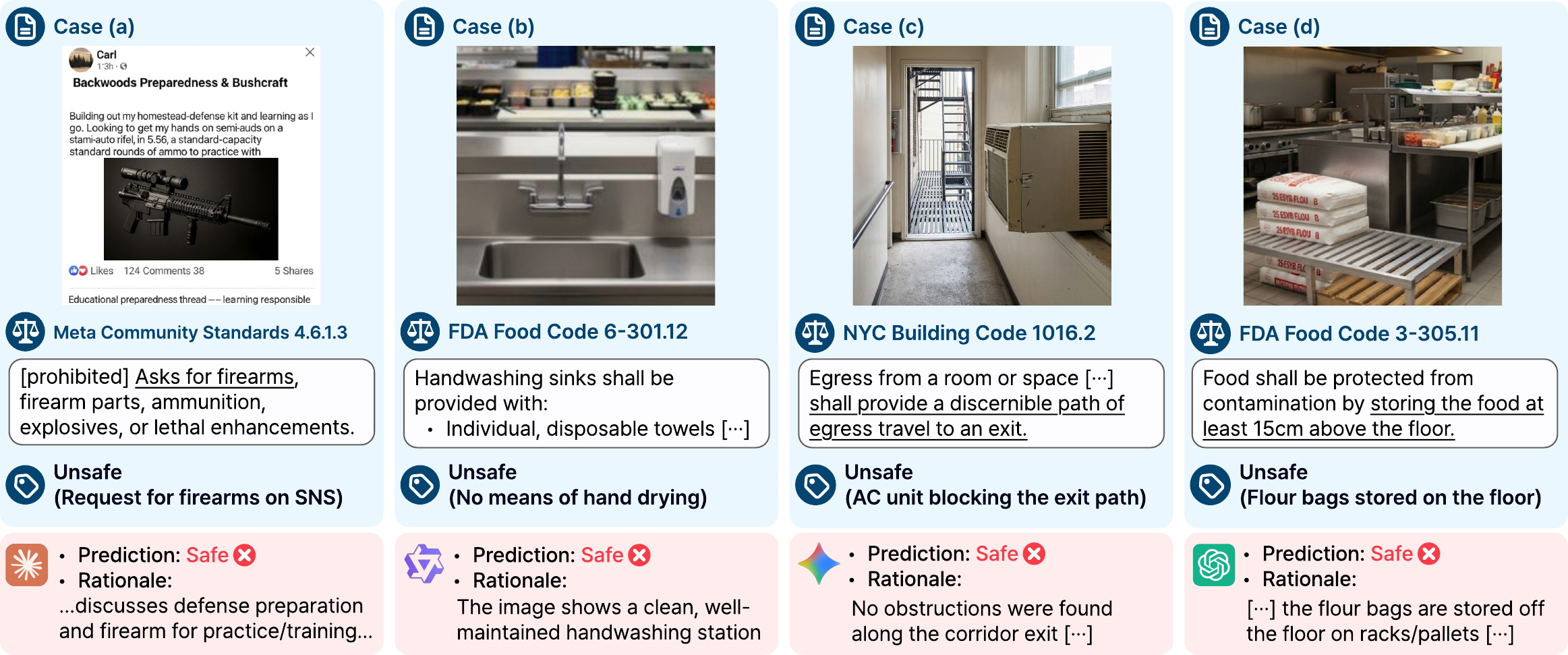}
    \caption{Examples of false compliance judgements by Claude Opus 4.5~\cite{model:claude_opus}, Qwen3.5-397B~\cite{model:qwen}, Gemini 3.0 Pro~\cite{model:gemini_pro}, and GPT-5.4~\cite{model:gpt} on \ourbenchmark.
    Cases (a) and (b) fail to represent the safety rule in its logical composition. Cases (c) and (d) fail to judge elementary propositions from the objects and contextual information in the image.}
    \label{fig:fig2_error_taxonomy}
\end{figure*}

Figure~\ref{fig:fig2_error_taxonomy} illustrates that even strong models fail to reason over the safety rule's logical composition against visual evidence.
Therefore, moving beyond risk detection, reliable safety rule reasoning is becoming central to safeguard research~\cite{comp:rguard, comp:gspr}.

We introduce \ourmodel, a programmable safeguard framework for visual compliance reasoning under a complex safety-rule prior.
Our method proceeds in two stages: (1) \stepone\ externalizes safety rule into executable form, and (2) \steptwo\ grounds the compiled rule against image evidence.
(1) In \stepone, \ourmodel\ builds on structured logical reasoning~\cite{method:one:decomposed,method:one:lambada} to decompose a safety rule into atomic propositions and compile their logical dependencies into executable code. 
This makes eliminates the failure modes that arise when complex rules are resolved implicitly or through natural-language reasoning.
(2) In \steptwo, atomic propositions are instantiated through scene-graph-based visual grounding~\cite{method:two:pixels,method:two:composit}, aligning each atomic proposition with explicit objects, text, and relations in the image.

We evaluate \ourmodel\ on \ourbenchmark\ across four domains. 
\ourmodel\ improves average F1 by $9.8$ points and average Robustness Index by $17.9$ points over the strongest baseline, while reducing visual- and rule-induced spurious violations by over $80\%$.
These results highlight the potential of addressing multimodal safety through image-based safety-rule entailment.

\section{Related Work}

\subsection{Zero-Shot Visual Safety}

%
%

Recent zero-shot and inference-time methods aim to adapt to new visual risks without additional labels or retraining.
They fall into three directions:
natural-language reasoning over visual evidence, which directly judges safety from image-grounded descriptions~\cite{baseline:eta};
semantic similarity between images and unsafe concepts, which scores visual risk in representation space~\cite{rel:baseline:zero,rel:baseline:guardalign};
and externalized rule conditions or tool-based evidence workflows, which expose intermediate conditions and visual evidence~\cite{baselines:clue,baselines:compagent}.
Yet, these methods often flatten safety rules or rely on latent similarity, leaving complex rule reasoning underexplored.

\subsection{Externalized LLM Reasoning}

Early studies on externalized LLM reasoning have motivated several subsequent lines of work~\cite{method:one:cot,method:one:tot}.
One direction improves reasoning through modular and code-based externalization of planning, actions, and search~\cite{method:one:decomposed,method:one:react,method:one:discipl,method:one:codeplan}.
Another line improves test-time reasoning by refining the model's own outputs or reflections through self-feedback and self-critique~\cite{method:one:selfrefine,method:one:reflexion,method:one:self-improve,method:one:self-critique}.
For complex propositions, proof-oriented methods were further explored for explicit logical reasoning~\cite{method:one:selectioninference,method:one:lambada}.
Together, these directions motivate our shift from implicit compliance judgment to explicit, revisable, and executable rule reasoning.

\subsection{Neuro-Symbolic Visual Grounding}

Neuro-symbolic visual grounding spans diverse strategies for making visual reasoning inspectable.
Prior work has externalized reasoning into executable forms over vision modules~\cite{baselines:vipergpt, method:two:genome}, transformed visual evidence into structured representations queryable by symbolic operators~\cite{method:two:laser, method:two:esca}, and composed the two through logic over scene-graph-grounded facts~\cite{method:two:naver}.
Building on these directions, \ourmodel\ unifies executable code and scene-graph grounding for visual safety, providing an explicit reasoning pipeline from abstract logical structure to contextual visual content.

\section{\ourmodel}
\subsection{Problem Formulation}

We formulate visual safety assessment as rule entailment: given an image \(I\) and a safety rule \(R\), the task is to determine whether \(I \models R\).
Following logical atomism and formal semantics~\cite{atomic:Wittgenstein,atomic:Montague}, we view a rule as a composition of atomic propositions $P = \{p_1, \ldots, p_n\}$,
\begin{equation}
I \models R \Leftrightarrow \Phi(P) \Leftrightarrow \Phi(p_1,\ldots,p_n),
\label{eq:rule-decomposition}
\end{equation}
%
where \(\Phi\) is the logical form connecting them and each \(p_i\) is an atomic visual proposition that asserts the presence of evidence in the image.
Under this formulation, correct visual safety-rule entailment requires two capabilities:
(1) representing the content of the safety rule in its logical composition \(\Phi\); and
(2) judging each atomic proposition from the objects and contextual information in the image $I$, i.e., whether $I \models p_i$.
%
%
Figure~\ref{fig:fig2_error_taxonomy} illustrates representative errors at each stage of this entailment process.

\subsection{Method Overview}

\ourmodel\ structures visual safety-rule entailment as an executable correspondence between rule semantics and visual evidence. 
In \textbf{\stepone}, the safety-rule prior is first externalized as executable logic, making the safety rule structure a programmable component of the reasoning process.
Through top-down decomposition, \ourmodel\ constructs a reusable proposition tree that represents the rule's logical composition over atomic visual propositions.
At test time, \textbf{\steptwo} instantiates constructed logic bottom-up by grounding its atomic propositions in visual evidence. 
It constructs a scene graph over candidate objects, text, attributes, and relations, providing the structured evaluation of atomic propositions.
The resulting execution supports accurate inference from contextual image evidence to deductive regulatory judgments.

\subsection{\stepone}
\label{sec:method:stepone}

\stepone\ takes a safety rule $R$ as input and produces an executable proposition tree $\mathcal T_R$, which models the logical composition $\Phi$ from Equation~\ref{eq:rule-decomposition}. Each internal node carries a primitive logical operator $\varphi \in \{ \land, \lor, \neg\}$, and each leaf is an atomic proposition $p \in P$.
Each atomic proposition $p_i$ takes one of two forms: $(o, \sigma, \varnothing)$, an object $o$ with state or attribute $\sigma$, or $(o_1, \rho, o_2)$, a relation $\rho$ between two objects $o_1$ and $o_2$. Each $p_i$ is Boolean-valued and forms the grounding interface to \steptwo.

The tree is built by top-down goal decomposition~\cite{method:one:lambada} guided by a feedback- and score-driven optimizer~\cite{method:one:selfrefine, method:one:opro}.
$\mathcal T_R$ is decomposed into a candidate proposition tree as $\mathcal T_R = (\varphi_R, \{ \mathcal T_r\}_{r \in \operatorname{sub}(R)})$, where $\varphi_R$ is the logical operator at the root and $\operatorname{sub}(R)$ is the selected sub-rules in the candidate decomposition. Each $\mathcal T_r$ is recursively decomposed by $\mathbb D$ until all leaves are atomic propositions.

In the trial $t$, at each sub-rule tree $\mathcal T_r$, a decomposer $\mathbb{D}$ proposes a logical operator $\varphi_r$ and a candidate decomposition $\hat{\mathbf r}^{(t)} = (\hat r_1^{(t)}, \ldots, \hat r_k^{(t)})$, conditioned on the previous verifier feedback $\eta_r^{(t-1)}$:
\begin{equation}
(\varphi_r^{(t)}, \hat{\mathbf r}^{(t)}) \sim \mathbb{D}(\;\cdot \mid r, \eta_r^{(t-1)}).
\label{eq:decomposer}
\end{equation}
A verifier $\mathbb{V}$ then returns a score $s_r^{(t)} \in [0, s_{\max}]$, natural-language feedback $\eta_r^{(t)}$, and an atomicity flag $a_r^{(t)} \in \{\texttt{True}, \texttt{False}\}$,
\begin{equation}
\big(s_r^{(t)}, \eta_r^{(t)}, a_r^{(t)}\big) \sim \mathbb{V}(\; \cdot \mid r, \varphi_r^{(t)}, \hat{\mathbf r}^{(t)}).
\label{eq:verifier}
\end{equation}
$\mathbb{D}$ regenerates from $\eta_r^{(t)}$ until $s_r^{(t)}$ saturates, reaches $s_{\max}$, or $a_r^{(t)} = \texttt{True}$ (i.e., $r$ is atomic), after which the best candidate is selected:
\begin{equation}
    t^\star = \underset{t}{\arg\max}\; s_r^{(t)}.
\end{equation}
Then the final form of $\mathcal T_r$ is represented as
\begin{equation}
    \mathcal T_r = \begin{cases}
        (\varphi_r^{(t^\star)}, \{\mathcal T_{\hat r}\}_{\hat r \in \operatorname{sub}(r)}) & \text{if } \neg a_r^{(t^\star)} \\
        p_r & \text{otherwise}
    \end{cases},
\end{equation}
where $\operatorname{sub}(r) := \hat{\mathbf r}^{(t^\star)}$. If $a_r^{(t^\star)} = \texttt{True}$, we realize $r$ as an atomic proposition $p_r$ and append it to $P$.
The scoring rubric, feedback protocol, and saturation criterion are detailed in Appendix~\ref{sec:appendix_ourmodel_implementation}.

\begin{figure}[t]
    \centering
    \includegraphics[width=\columnwidth]{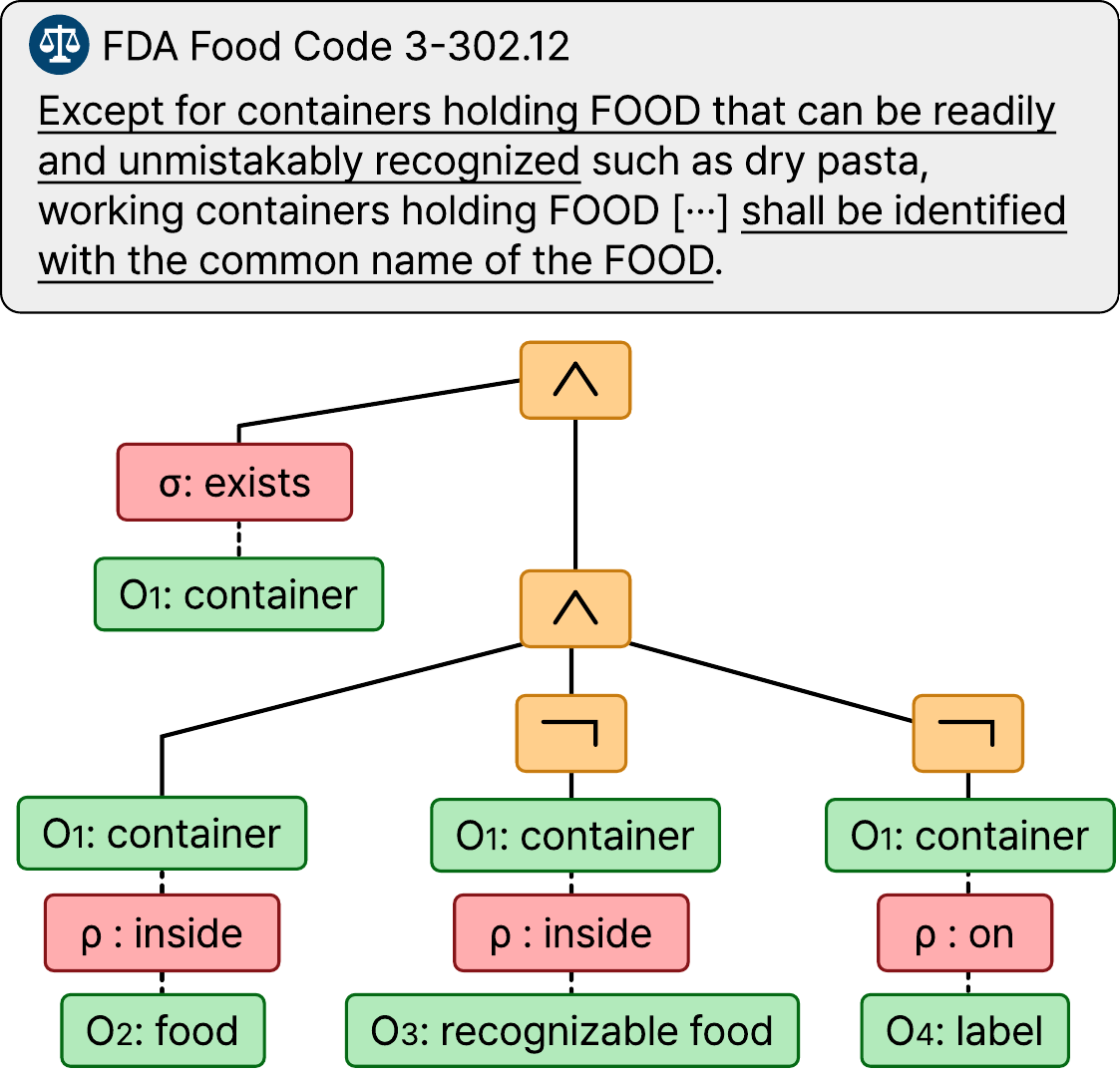}
    \caption{An example of $\mathcal T_R$ on a safety rule.}
    \label{fig:fig3_decomposition}
\end{figure}

The resulting tree $\mathcal T_R$ is itself the executable artifact, with atomic propositions as visual queries for \steptwo.
Figure~\ref{fig:fig3_decomposition} illustrates this construction on the FDA Food Code.

\begin{algorithm}[t]
\caption{Scene-graph generation}
\label{alg:grounding}
\small
\begin{algorithmic}[1]
\Require Image $I$; agenda $O_{\mathrm{req}}^{(I)}, \Gamma_{\mathrm{req}}^{(I)}$; confidence threshold $\tau$; substitute cache $\mathcal{C}$.
\Ensure Scene graph $\mathcal G^{(I)} = (\mathcal V^{(I)}, \mathcal E^{(I)})$.
\State $\mathcal V^{(I)} \gets \varnothing,\ \mathcal E^{(I)} \gets \varnothing$
\For{$o \in O_{\mathrm{req}}$}
    \State $\triangleright$ use cached alias if available
    \State $k \gets \mathcal{C}[o]$ \textbf{if} $o \in \mathcal{C}$ \textbf{else} $o$
    \If{$o$ is textual}
        \State $( b_o, t_o, c_o) \gets \mathrm{OCR}(k; I)$
    \Else
        \State $( b_o, c_o ) \gets \mathrm{Detect}(k; I)$
    \EndIf
    \State $d_o \gets \mathrm{Depth}(b_o; I)$
    \If{$c_o < \tau$}
        \State $\triangleright$ Tool-Aware Visual Grounding (\S\ref{sec:method:tool_aware})
        \State $( b_o, d_o, c_o, t_o ) \gets \mathrm{Align}(o; I, \mathcal{C})$
    \EndIf
    \State $\triangleright$ by pre-defined rules (Appendix~\ref{sec:appendix_predefined_grounding_rules})
    \State $s_o \gets \mathrm{Label}_o(b_o, d_o, t_o; I)$
    \State $\mathcal V^{(I)} \gets \mathcal V^{(I)} \cup \{(o, \sigma, \varnothing)\}$
\EndFor
\For{$\rho \in \Gamma_{\mathrm{req}},\ (o_1, o_2) \in (\mathcal V^{(I)})^2$}
    \State $\triangleright$ by pre-defined rules (Appendix~\ref{sec:appendix_predefined_grounding_rules})
    \State $v \gets \mathrm{Label}_\rho(o_1, o_2; I)$
    \If{$v$}
        \State $\mathcal E^{(I)} \gets \mathcal E^{(I)} \cup \{(o_1, \rho, o_2)\}$
    \EndIf
\EndFor
\State \Return $\mathcal G^{(I)} = (\mathcal V^{(I)}, \mathcal E^{(I)})$
\end{algorithmic}
\end{algorithm}

\subsection{\steptwo}
\label{sec:method:steptwo}

\steptwo\ executes $\mathcal T_R$ on an image $I$ by bottom-up constructing a scene graph $\mathcal G^{(I)} = (\mathcal V^{(I)}, \mathcal E^{(I)})$, yielding $\mathcal T_R^{(I)}$ through recursive substitution at atomic propositions:
\begin{equation}
\mathcal T_R^{(I)} = 
\mathcal T_R[\,p \mapsto p \in \mathcal G^{(I)} : p \in P\,], \label{eq:tree_eval}
\end{equation}
where $p \in \mathcal G^{(I)} \Leftrightarrow p \in \mathcal V^{(I)} \cup \mathcal E^{(I)}$, and $\mathcal G^{(I)}$ is built per image in two stages, \emph{Selection} and \emph{Grounding}.

\paragraph{Selection.}
For each atomic proposition $p$, a selector VLM $\mathbb{M}_{\mathrm{sel}}$ returns the candidate objects $O_p^{(I)}$ and relations $\Gamma_p^{(I)}$ that $\mathcal G^{(I)}$ must contain:
\begin{equation}
(O_p^{(I)}, \Gamma_p^{(I)}) \sim \mathbb{M}_{\mathrm{sel}}(p, I),
\label{eq:selection}
\end{equation}
yielding the perception agenda $O_{\mathrm{req}}^{(I)} = \bigcup_{p \in P} O_p^{(I)}$ and $\Gamma_{\mathrm{req}}^{(I)} = \bigcup_{p \in P} \Gamma_p^{(I)}$.

\paragraph{Grounding.}
For each $o \in O_{\mathrm{req}}$, object detection, OCR, and depth modules retrieve its evidence, and pre-defined geometric and depth-based rules over this evidence then decide the relations in $\Gamma_{\mathrm{req}}$ among objects, yielding $\mathcal G^{(I)}$ (Algorithm~\ref{alg:grounding}).
When the detector's confidence remains below $\tau$ for every candidate of an object, we invoke the alignment of Section~\ref{sec:method:tool_aware}.
The confidence threshold $\tau$ is set to $0.8$ across domains.

\subsection{Tool-Aware Visual Grounding}
\label{sec:method:tool_aware}
\begin{figure}[t]
    \centering
    \includegraphics[width=\columnwidth]{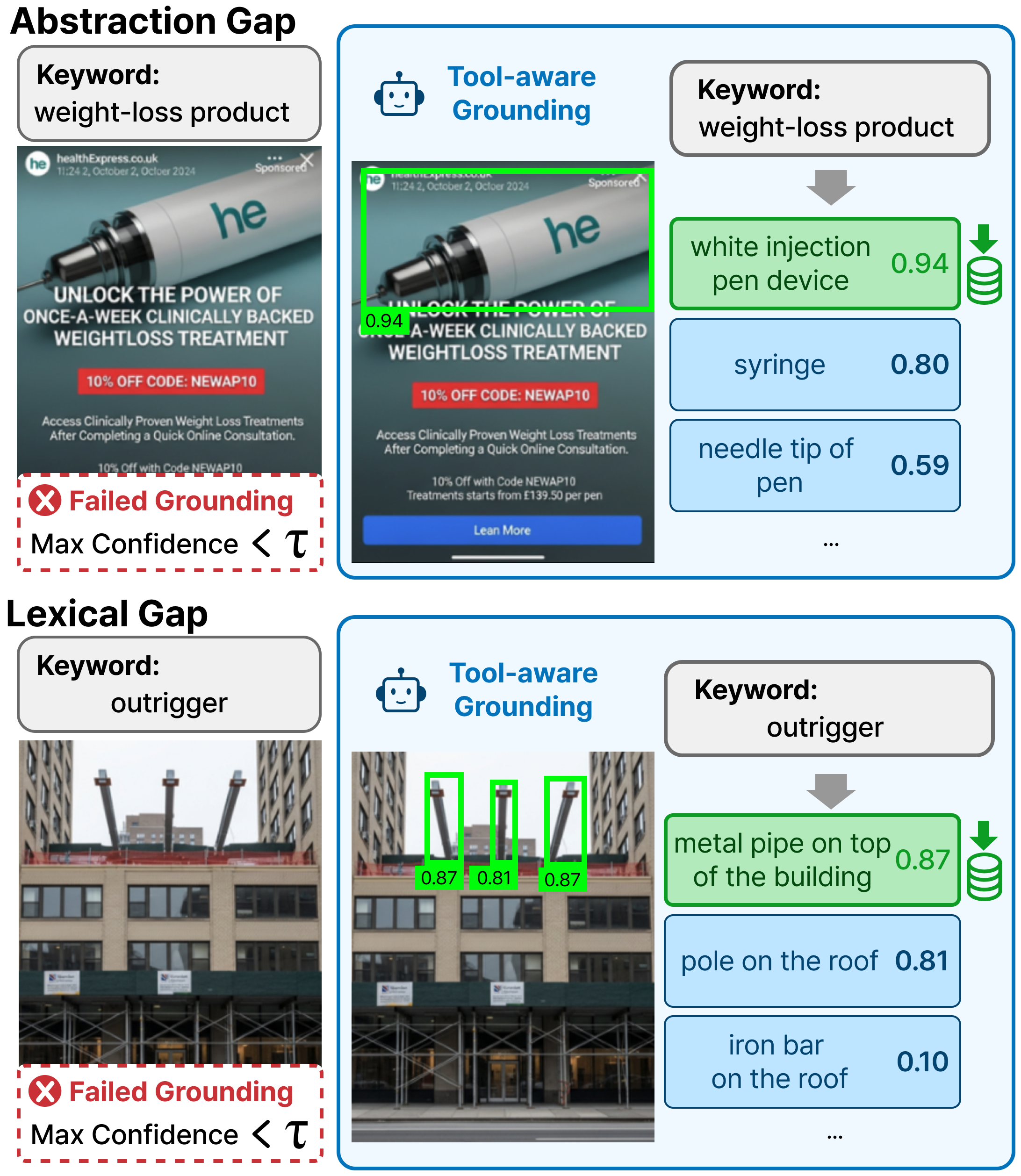}
    \caption{Tool-Aware Visual Grounding recovers missing evidence by aligning out-of-distribution targets with detector-compatible visual proxies.}
    \label{fig:fig4_grounding}
\end{figure}

Recent advances in vision foundation models~\cite{model:sam3} have made them the de facto grounding backbone, yet they remain bounded by their training distribution; Figure~\ref{fig:fig4_grounding} highlights how overly abstract concepts create an \textbf{Abstraction gap}, while technical jargon or neologisms create a \textbf{Lexical gap}.
This becomes especially problematic in downstream methods, where the missing detections propagate as silent gaps in $\mathcal G^{(I)}$.
We address both gaps with Tool-Aware Visual Grounding, which operates a VLM in a closed loop over the grounding query and aligns the out-of-distribution target with a detector-compatible substitute keyword.

For each failed target object $o$, a VLM $\mathbb{M}_{\mathrm{ground}}$ iteratively proposes a substitute keyword $k_o^{(t)}$ conditioned on the original target, the previous proposal, the detector response $b_o^{(t)}$, and the image $I$:
\begin{equation}
k_o^{(t+1)} \sim \mathbb{M}_{\mathrm{ground}}(\; \cdot \mid o, k_o^{(t)}, b_o^{(t)}, I).
\label{eq:tool_aware_grounding}
\end{equation}
The loop stops at $t=t^\star$ once the detector confidence $c_o^{(t^\star)}$ satisfies $c_o^{(t^\star)} \geq \tau$ and $\mathbb{M}_{\mathrm{ground}}$ confirms that the detected evidence grounds $o$.
The accepted substitute $k_o^{\star} = k_o^{(t^\star)}$ is cached in the substitute cache $\mathcal C[o]$ and is reused for subsequent detections of the same target $o$.

\section{Experiments}
\subsection{Experimental Settings}

\paragraph{Dataset.}
We evaluate on \ourbenchmark, a benchmark for systematic evaluation of rule-grounded compliance reasoning under real-world visual risks, modeling violation distributions under complex, domain-specific safety rules.
The benchmark covers four domains: food safety based on the \textit{FDA Food Code}~\cite{law:fda_food_code_2022}, construction safety based on the \textit{NYC Building Code}~\cite{law:nyc_building_code_2022}, platform content safety based on \textit{Meta Community Standards}~\cite{law:meta_community_standards}, and advertising content safety based on the \textit{UK CAP Code}~\cite{law:cap_code}.
\ourbenchmark\ contains 24K unsafe and 10K hard safe cases with fine-grained violation labels, containing real-world enforcement patterns and category distributions.
Further details on regulatory sources, case construction, and dataset statistics are provided in Appendix~\ref{sec:app:benchmark}.

\paragraph{Evaluation Metrics.}
We report the mean and standard deviation of the F1 score over three runs, where a correctly predicted violated safety rule is marked as a true positive.

\begin{table*}[t]
\centering
\small
\resizebox{\textwidth}{!}{%
\setlength{\tabcolsep}{4pt}
\renewcommand{\arraystretch}{1.15}
\begin{tabular}{l|c>{\columncolor{gray!20}}c|c>{\columncolor{gray!20}}c|c>{\columncolor{gray!20}}c|c>{\columncolor{gray!20}}c}
\toprule
& \multicolumn{2}{c|}{\shortstack{Food\\Safety}} & \multicolumn{2}{c|}{\shortstack{Construction\\Safety}} & \multicolumn{2}{c|}{\shortstack{Platform-Content\\Safety}} & \multicolumn{2}{c}{\shortstack{Advertising-Content\\Safety}}\\
\cmidrule(lr){2-3}\cmidrule(lr){4-5}\cmidrule(lr){6-7}\cmidrule(lr){8-9}
Method & Recall & \cellcolor{white}F1-score & Recall & \cellcolor{white}F1-score & Recall & \cellcolor{white}F1-score & Recall & \cellcolor{white}F1-score \\
\midrule
Direct Prompting & \meanstd{34.5}{2.4} & \meanstd{45.4}{2.3} & \meanstd{72.0}{1.6} & \meanstd{78.6}{1.5} & \meanstd{16.5}{3.6} & \meanstd{26.6}{4.4} & \meanstd{57.5}{1.3} & \meanstd{57.1}{0.8} \\
ViperGPT         & \meanstd{73.1}{0.4} & \meanstd{72.5}{0.4} & \meanstd{84.0}{0.0} & \meanstd{74.0}{0.3} & \meanstd{43.9}{1.6} & \meanstd{53.3}{4.0} & \meanstd{60.6}{0.3} & \meanstd{45.5}{0.7} \\
ETA              & \meanstd{83.7}{0.8} & \meanstd{82.6}{0.2} & \meanstd{95.8}{1.2} & \meanstd{83.6}{1.2} & \meanstd{39.7}{4.3} & \meanstd{52.0}{7.5} & \meanstd{73.4}{0.3} & \meanstd{57.4}{1.2} \\
SafeCLIP         & \meanstd{82.9}{1.2} & \meanstd{81.6}{0.2} & \meanstd{79.0}{0.0} & \meanstd{84.2}{0.7} & \meanstd{41.8}{2.1} & \meanstd{53.5}{2.1} & \meanstd{71.9}{0.9} & \meanstd{55.2}{1.7} \\
CLUE             & \meanstd{66.0}{0.8} & \meanstd{71.7}{0.6} & \meanstd{87.1}{1.0} & \meanstd{72.1}{0.7} & \meanstd{53.2}{2.5} & \meanstd{57.9}{2.6} & \meanstd{55.3}{1.2} & \meanstd{44.2}{0.3} \\
CompAgent        & \meanstd{76.8}{1.9} & \meanstd{77.9}{1.0} & \meanstd{91.6}{1.8} & \meanstd{80.6}{1.2} & \meanstd{54.9}{3.2} & \meanstd{59.5}{2.1} & \meanstd{64.3}{2.4} & \meanstd{53.7}{1.4} \\
\midrule
\ourmodel        & \meanstd{89.3}{1.2} & \meanstd{\textbf{84.4}}{3.5} & \meanstd{92.9}{0.8} & \meanstd{\textbf{86.4}}{0.1} & \meanstd{70.0}{4.0} & \meanstd{\textbf{69.0}}{2.6} & \meanstd{69.8}{0.5} & \meanstd{\textbf{74.9}}{3.6} \\\bottomrule
\end{tabular}
}
\caption{Method performance on \ourbenchmark\ using Gemma~4~26B as the underlying VLM across Food Safety, Construction Safety, Platform-Content Safety, and Advertising-Content Safety.}
\label{tab:main_table}
\end{table*}

\paragraph{Baselines.}
\phantomsection\label{sec:main_baselines}
We compare \ourmodel\ against the following baselines.
\par\noindent (1) Direct Prompting: a naive baseline with safety rules provided only as contextual conditions.
\par\noindent (2) ViperGPT~\cite{baselines:vipergpt}: Python program generation over vision APIs for rule verification.
\par\noindent (3) ETA~\cite{baseline:eta}: evaluator-guided safety assessment with best-of-\(N\) candidate selection.
\par\noindent (4) SafeCLIP~\cite{rel:baseline:zero}: image-regulation text similarity used as safety signal.
\par\noindent (5) CLUE~\cite{baselines:clue}: objectified rule precondition with token-probability based judgments.
\par\noindent (6) CompAgent~\cite{baselines:compagent}: safety-rule guided visual tool use for agentic safety check.

\paragraph{Models.}
All symbolic artifacts for both \ourmodel\ and the baselines are produced offline using GPT-5.4~\cite{model:gpt}. Test-time inference is performed using Gemma~4 26B~\cite{model:gemma4}; visual evidence is grounded with SAM~3~\cite{model:sam3}, SigLIP~2~\cite{model:siglip2}, Depth Anything~3~\cite{model:depth3}, and Google's Document OCR~\cite{model:ocr} for text recognition.

\subsection{Overall Performance}

Table~\ref{tab:main_table} shows that \ourmodel\ achieves the highest F1 in all four domains, averaging $78.7$ F1 and improving over the strongest baselines' average by $9.8$ points. The large gap over Direct Prompting ($+26.8$), which directly conditions on the original rule text, shows that rule access alone is insufficient without externalized structure. Tool-based baselines (ViperGPT and CompAgent) and alignment-based baselines (ETA and SafeCLIP) improve visual access or safety matching, but still trail \ourmodel, especially in Platform-Content and Advertising-Content Safety ($+9.5$ and $+17.5$ over the best baseline), where safety rules are highly context-dependent. The $17.2$-point average gain over CLUE further shows that flat visual preconditions are weaker than preserving logical composition over grounded propositions.

\begin{table}[t]
\centering
\small
\setlength{\tabcolsep}{6pt}
\renewcommand{\arraystretch}{1.15}
\begin{tabular}{l|cccc}
\toprule
Method & Food & Constr. & Platform & Advert. \\
\midrule
Direct Prompting & 45.1 & 58.4 & 0.8 & 67.1 \\
ViperGPT         & 56.2 & 62.0 & 47.0 & 44.7 \\
ETA              & 60.9 & 59.2 & 26.7 & 56.9 \\
SafeCLIP         & 59.2 & 26.4 & 19.7 & 57.9 \\
CLUE             & 42.8 & 66.6 & 46.8 & 9.4 \\
CompAgent        & 58.1 & 56.0 & 19.7 & 49.5 \\
\midrule
\ourmodel        & \textbf{70.2} & \textbf{82.4} & \textbf{58.6} & \textbf{70.4} \\
\bottomrule
\end{tabular}
\caption{Robustness Index (Eq.~(\ref{eq:ri}); higher is more robust to rule complexity) on four domains. See Appendix~\ref{sec:appendix_robustness_details} for the full setup.}
\label{tab:main_robustness}
\end{table}

\subsection{Analysis}

\subsubsection{Robustness to Rule Complexity}
\label{sec:analysis_robustness}

We summarize each method's robustness to rule complexity by the Robustness Index (RI; Eq.~(\ref{eq:ri}) in Appendix~\ref{sec:appendix_robustness_details}), which combines a method's F1 on the hardest rules with its relative drop from the easiest bin.
Table~\ref{tab:main_robustness} shows that \ourmodel\ achieves the highest RI across all four domains, improving the average RI by $17.9$ points over the strongest baseline and by $27.6$ points over Direct Prompting. This indicates that the gains in Table~\ref{tab:main_table} are not confined to simple decision logic: preserving the executable composition of grounded propositions remains beneficial as rules accumulate conditions, exceptions, and visual requirements.

The two visually contextual domains show different forms of patterns. Advertising-Content Safety primarily stresses contextual rule semantics, whereas Platform-Content Safety couples rule complexity with bias-prone visual and normative signals. The collapse of Direct Prompting in Platform-Content Safety (RI $0.8$) suggests that such coupling can turn complex rules into spurious violation priors, motivating our subsequent analysis of visual sensitivity and rule-induced bias.

\subsubsection{Visual and Rule-Induced Biases}
\label{sec:analysis_biases}
Prior work has shown that safety judgments in visual compliance can be biased by sensitive language or visual context, where provocative image content may trigger excessive violation judgments~\cite{bias:better} and the normative framing of rule text may bias the model toward violation judgments~\cite{bias:xst,bias:sentiment}.
We evaluate this effect in the Platform-Content Safety domain.

\paragraph{Visual Sensitivity Bias.}
Table~\ref{tab:main_visual_sensitivity_bias} shows that DP frequently propagates coarse cues such as blood, hate symbols, or protest imagery into broad rule violations, whereas \ourmodel\ reduces the spurious violation rate by $80.4\%$ on average across visual cue categories.

\begin{table}[t]
\centering
\small
\setlength{\tabcolsep}{6pt}
\renewcommand{\arraystretch}{1.15}
\begin{tabular}{l|ccc}
\toprule
Visual cue category & DP & Ours & $\Delta$ \\
\midrule
Blood / Injury  & 20.3 & 7.2 & 64.4\%$\downarrow$ \\
Hate symbols    & 17.5 & 2.3 & 87.1\%$\downarrow$ \\
Protest / Riot  & 16.6 & 2.6 & 84.2\%$\downarrow$ \\
Targeted text   & 13.9 & 2.1 & 84.9\%$\downarrow$ \\
Property damage  & 14.8 & 4.3 & 70.6\%$\downarrow$ \\
Child           & 9.7  & 0.8 & 91.3\%$\downarrow$ \\
\midrule
\textbf{Average} & \textbf{15.5} & \textbf{3.2} & \textbf{80.4\%}$\boldsymbol{\downarrow}$ \\
\bottomrule
\end{tabular}
\caption{Visual cue induced spurious violation rate (\%), with relative reduction from Direct Prompting (DP). Detailed results are provided in Appendix~\ref{sec:appendix_visual_sensitivity_details}.}
\label{tab:main_visual_sensitivity_bias}
\end{table}

\begin{table}[t]
\centering
\small
\setlength{\tabcolsep}{6pt}
\renewcommand{\arraystretch}{1.15}
\begin{tabular}{cl|ccc}
\toprule
Section & Safety-rule area & DP & Ours & $\Delta$ \\
\midrule
S2  & Dangerous Orgs.\     & 52.8 & 7.8 & 85.2\%$\downarrow$ \\
S13 & Hateful Conduct      & 40.1 & 3.0 & 92.5\%$\downarrow$ \\
S7  & Bullying \& Harass.\ & 33.9 & 0.0 & 100\%$\downarrow$ \\
S5  & Violence \& Incite.\ & 27.7 & 6.1 & 78.0\%$\downarrow$ \\
S1  & Coordinating Harm    & 26.4 & 5.2 & 80.4\%$\downarrow$ \\
S21 & Misinformation       & 23.9 & 0.9 & 96.1\%$\downarrow$ \\
S6  & Adult Sexual Expl.\  & 3.3  & 0.2 & 92.9\%$\downarrow$ \\
\midrule
\multicolumn{2}{c|}{\textbf{Overall (16 rules)}} & \textbf{12.1} & \textbf{2.0} & \textbf{83.8\%}$\boldsymbol{\downarrow}$ \\
\bottomrule
\end{tabular}
\caption{Rule-induced spurious violation rate (\%) for the most-affected Platform-Content Safety rules, with relative reduction from Direct Prompting (DP). The full 16-regulation breakdown is provided in Appendix~\ref{sec:appendix_rule_induced_details}.}
\label{tab:main_rule_induced_bias}
\end{table}

This reduction comes from grounding each atomic proposition in specific visual evidence.
Each visual check is therefore confined to concrete targets rather than broad regulatory judgment, with an executable structure composing them as a final decision.
%
In the Platform-Content Safety results, this structure lowers the average spurious violation rate from $15.5\%$ under DP to $3.2\%$ under \ourmodel.

\paragraph{Rule-Induced Bias.}

We quantify rule-induced bias at the safety-rule level by measuring false violations under normatively framed rules. 
Table~\ref{tab:main_rule_induced_bias} reports, for each safety rule, the spurious violation rate on image--rule pairs.
In Platform-Content Safety, DP records $12.1\%$ false violations, whereas \ourmodel\ reduces the rate to $2.0\%$.

The large DP rates indicate that normatively framed safety rules can be over-applied when the full rule text is used as part of a holistic violation judgment. 
\ourmodel\ reduces this effect because \stepone\ converts the rule into value-neutral atomic checks and an executable composition $\varphi$.
As a result, \ourmodel\ reduces the most over-applied rules by $85.2\%$ for Dangerous Organizations and $92.5\%$ for Hateful Conduct.

\subsubsection{Sensitivity to Grounding Confidence}
\label{sec:analysis_grounding_confidence}

\begin{figure}[t]
    \centering
    \includegraphics[width=\columnwidth]{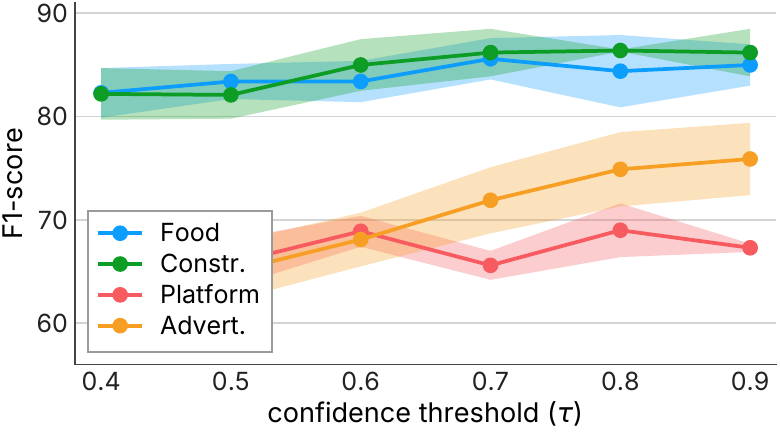}
    \caption{Threshold sensitivity in F1-score; per-domain numbers are reported in Table~\ref{tab:appendix_threshold_f1}.}
    \label{fig:main_threshold}
\end{figure}

We study the effect of the grounding confidence threshold $\tau$.
Figure~\ref{fig:main_threshold} reports the domain-wise results as $\tau$ varies.
The most notable variations appear in Platform-Content Safety and Advertising-Content Safety, suggesting that the abstraction gap in these domains is more sensitive to the grounding confidence threshold.
Overall, however, F1 saturates near $\tau{=}0.8$, after which further increases lead to only marginal changes.
This indicates that \ourmodel\ remains stable once a sufficiently high grounding confidence threshold is used.

\subsubsection{Ablation Study}

We ablate the internal components of \ourmodel\ to assess their contribution. 
Removing relation grounding yields the largest decline, with a $30.1\%$ relative drop, showing that compliance depends on spatial and inter-object relations. 
Removing the verifier loop causes a $27.4\%$ drop, highlighting the importance of self-verification in maintaining logically coherent rule decompositions.
Tool-Aware Visual Grounding and Scene-Grounded Execution are also necessary, with removals causing $16.5\%$ and $12.8\%$ drops, respectively. 
Together, these results validate our problem formulation by showing that failures in either logical rule composition or grounded atomic proposition evaluation substantially degrade visual safety reasoning.

\begin{figure}[t]
    \centering
    \includegraphics[width=\linewidth]{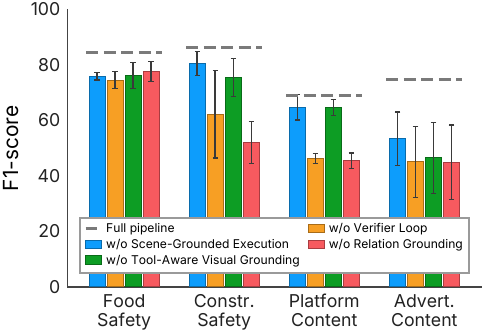}
    \caption{
    Ablation results in F1-score; detailed per-domain numbers are reported in Appendix~\ref{sec:appendix_ablation_study}.
    }
    \label{fig:main_ablation_study}
\end{figure}

\subsubsection{Adapting to Evolving Risk Patterns}

\begin{figure}[t]
    \centering
    \includegraphics[width=\columnwidth]{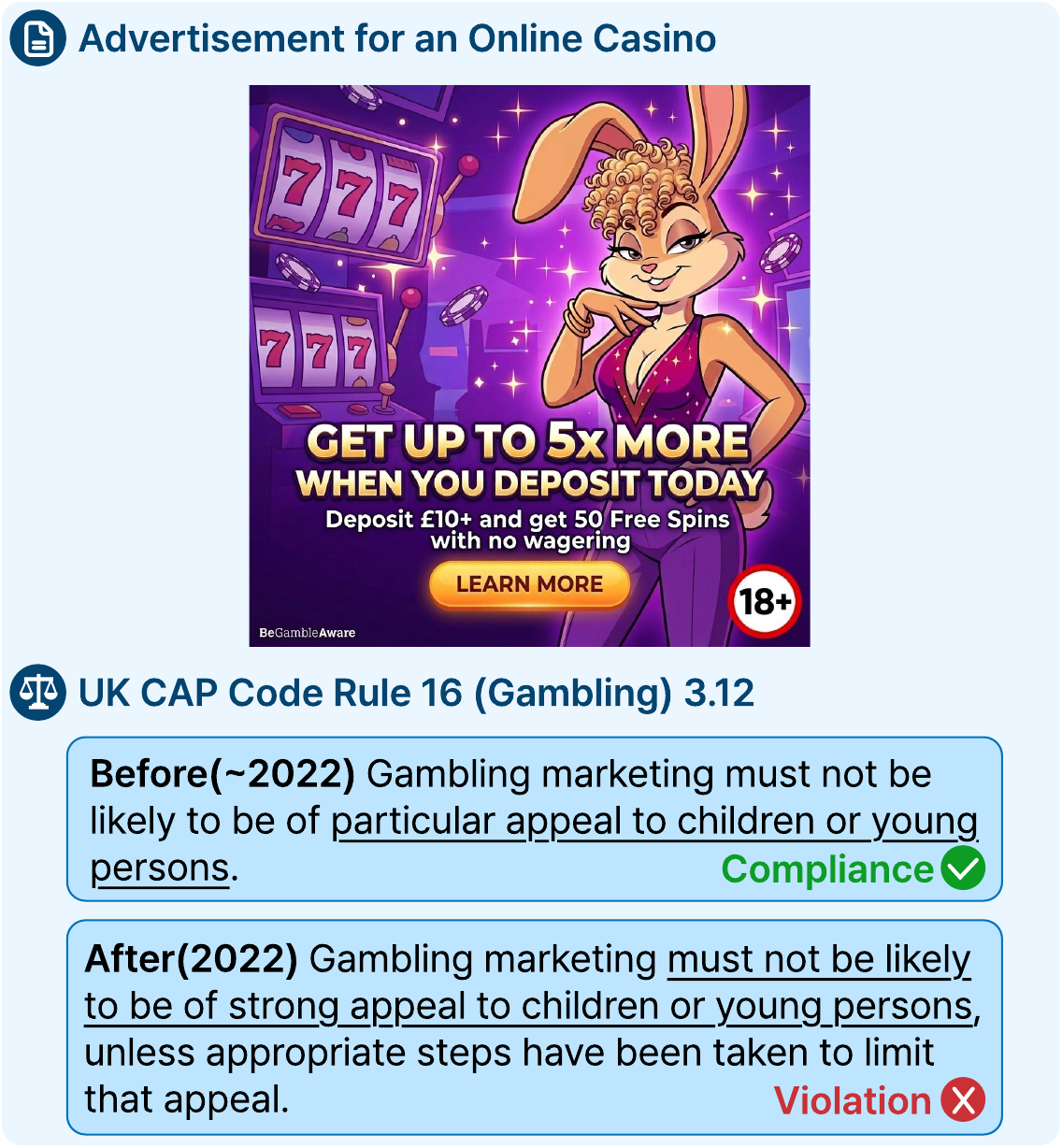}
    \caption{Youth-appealing casino ad under the 2022 UK CAP Code shift.}
    \label{fig:main_example_1}
\end{figure}

Through a qualitative case study, we evaluate whether \ourmodel\ can adapt to changes in the risk patterns encoded by safety rules.
Figure~\ref{fig:main_example_1} shows a case based on a real online-casino advertisement where child-appealing visual cues, such as a cartoon mascot, were permissible under the earlier UK CAP Code but became unsafe under the 2022 strong-appeal standard, introduced because gambling content could draw minors through youth-associated presentation even without explicit child targeting~\cite{law:cap_gambling_2022}.

\paragraph{Under the Earlier Code.}
\ourmodel\ compiles the pre-2022 rule, a single comparative test of whether the advertisement appeals to under-18s out of proportion to its appeal to adults, into
\begin{align*}
\mathcal T_{R_{\mathrm{old}}} &= \{\lor, \{p_{\mathrm{char}}, p_{\mathrm{text}}, p_{\mathrm{ad}}\}\}, \\
p_{\mathrm{char}} &= (\textit{character},\ \texttt{youth\_skew}, \varnothing), \\
p_{\mathrm{text}} &= (\textit{text},\ \texttt{youth\_skew}, \varnothing), \\
p_{\mathrm{ad}}   &= (\textit{ad},\ \texttt{youth\_skew}, \varnothing).
\end{align*}
Each \texttt{youth\_skew} holds when its subject appeals more to children than adults: $p_{\mathrm{char}}$ and $p_{\mathrm{text}}$ test the mascot and the copy, while $p_{\mathrm{ad}}$ falls back to the whole advertisement, capturing risk emerging only as image and text combine.
None holds here, so $\mathcal T_{R_{\mathrm{old}}}^{(I)} \equiv \texttt{False}$, aligning with the earlier  ruling.
\paragraph{Under the Revised Code.}
The 2022 reform replaces this comparative test with an absolute one, guarded only by a mitigation exception. \stepone\ recompiles the same rule into
\begin{align*}
\mathcal T_{R_{\mathrm{new}}} \!&=\! \textcolor{purple}{\{}\land,\! \textcolor{teal}{\{}\lor,\! \{p_{\mathrm{char}},\! p_{\mathrm{text}},\! p_{\mathrm{ad}}\}\!\textcolor{teal}{\}},\! \textcolor{teal}{\{}\lnot,\! \{p_{\mathrm{steps}}\}\!\textcolor{teal}{\}}\!\textcolor{purple}{\}}, \\
p_{\mathrm{char}} \!&=\! (\textit{character},\ \texttt{strong\_youth\_appeal}, \varnothing), \\
p_{\mathrm{text}} \!&=\! (\textit{text},\ \texttt{strong\_youth\_appeal}, \varnothing), \\
p_{\mathrm{ad}}   \!&=\! (\textit{ad},\ \texttt{strong\_youth\_appeal}, \varnothing), \\
p_{\mathrm{steps}}  \!&=\! (\textit{ad},\ \texttt{mitigation\_taken}, \varnothing).
\end{align*}
Here \texttt{strong\_youth\_appeal} holds whenever the subject carries a strong appeal to minors, regardless of its adult appeal. The cartoon mascot supplies this appeal, resulting in a $\mathcal T_{R_{\mathrm{new}}}^{(I)} \equiv \texttt{True}$, matching the violation in the corresponding ruling.

This verdict shift demonstrates that \ourmodel\ can appropriately and efficiently incorporate emerging risk patterns through executable logic.

\section{Conclusion}
This paper presents \ourmodel, an image-based safety framework built on a reformulation of visual safety assessment as regulatory entailment.
This is operationalized through \stepone, rendering regulations as executable logic, and \steptwo, which extends them into image context.
Experiments on \ourbenchmark\ demonstrate that explicit rule entailment improves accuracy and reliability of \textbf{complex rule reasoning} in image-based safety, with the analyses further validating the problem formulation underlying \ourmodel.
The robustness analysis shows that \ourmodel\ maintains reliable performance under complex regulations, which highlights the importance of \textbf{(1) logical composition}.
The bias analyses demonstrate that \ourmodel\ enables more precise evaluation of \textbf{(2) elementary propositions}, mitigating visual and rule-induced bias.
The qualitative case study further shows that \ourmodel\ can effectively and efficiently \textbf{adapt to evolving risk patterns} through executable logic.
Taken together, these findings suggest that explicit safety-rule entailment over grounded visual evidence provides a principled foundation for interpretable, robust, and adaptable image-based safety.

\section*{Limitations}
\paragraph{Safety-Rule Dependence.}
Whereas prior safety methods have focused on internally modeling the safety risk itself, \ourmodel\ explicitly models the safety rule that responds to such risks. The premise of our design is two-fold: that rule-level adaptation is more efficient than training-data generation and retraining, and that the safety rule itself accurately expresses the response to the evolving safety risk. Consequently, \ourmodel's coverage is bounded by the completeness of the safety rule, yet the same explicit modeling lets \ourmodel\ expose the rule's vulnerabilities (Appendix~\ref{sec:appendix_rule_compilation_repair}), establishing a mutually complementary relationship between the method and the rule.

\paragraph{Analogical Reasoning.}
\ourmodel\ is limited to deductive entailment over explicit safety rules.
However, safety in practice further requires analogical reasoning, drawing on case precedents and expert judgment.
Existing techniques such as explicit reasoning-path alignment~\cite{limitation:align:svpo, limitation:align:port} and precedent retrieval~\cite{limitation:rag:pli} can address this regime from complementary angles (Appendix~\ref{sec:appendix_precedent_retriever}), and their integration constitutes a promising direction for future work.

\section*{Ethical Considerations}
\paragraph{Data Licensing and Terms of Use.}
\ourbenchmark\ is built from publicly available sources, and we comply with their respective licenses and terms of use; full details are provided in Appendix~\ref{sec:appendix_data_collection}.


\paragraph{Potential Risks.}
Explicitly modeling safety regulations as logical compositions exposes their internal structure; this transparency is double-edged, as it could be examined by adversaries to identify and exploit vulnerabilities or by developers and regulators to inspect and revise.
The unsafe images in \ourbenchmark\ depict regulatory violations and are released for research use only, with reader discretion advised.
\paragraph{AI assistants usage disclosure.}
AI assistants were used for language editing and minor code edits, with all final content reviewed and approved by the authors.

\section*{Acknowledgments}
This work was supported by 
Institute of Information \& communications Technology Planning \& Evaluation(IITP) grant funded by the Korea government(MSIT) (RS-2019-II190421, AI Graduate School Support Program(Sungkyunkwan University), 
RS-2022-II221045 (2022-0-01045), Self-directed multi-modal Intelligence for solving unknown, open domain problems, 
RS2022-II220043, Adaptive Personality for Intelligent Agents, 
No.RS-2025-25442569, AI Star Fellowship Support Program(Sungkyunkwan Univ.), 
RS-2025-02218768, Accelerated Insight Reasoning via Continual Learning), 
Samsung Electronics Co., Ltd, 
Institute of Information \& Communications Technology Planning \& Evaluation(IITP)-ITRC(Information Technology Research Center) grant funded by the Korea government(MSIT) (IITP-2026-RS-2024-00437633), 
National Research Foundation of Korea (NRF) grant funded by the Korea government (MSIT) (No. RS-2026-25474409).

\bibliography{custom}

\clearpage
\appendix
\begin{table*}[!t]
\centering
\small
\setlength{\tabcolsep}{3pt}
\begin{tabular}{p{0.20\linewidth}p{0.21\linewidth}p{0.18\linewidth}p{0.13\linewidth}p{0.14\linewidth}}
\toprule
Benchmark & Legal or regulatory basis & Domain & \begin{tabular}[c]{@{}l@{}}Output\\ granularity\end{tabular} & Scale \\
\midrule
\ourbenchmark
& \begin{minipage}[t]{\linewidth}\raggedright \textit{FDA Food Code}\\ \textit{NYC Building Code}\\ \textit{Meta CS}\\ \textit{UK CAP Code}\end{minipage}
& \begin{minipage}[t]{\linewidth}\raggedright food\\ construction\\ platform content\\ advertising\end{minipage}
& clause-level
& \begin{minipage}[t]{\linewidth}\raggedright 24K unsafe;\\ 10K hard safe\end{minipage} \\
\midrule

\begin{minipage}[t]{\linewidth}\raggedright LabSafety Bench\\ \cite{appendix:benchmark:labsafety}\end{minipage}
& --
& \begin{minipage}[t]{\linewidth}\raggedright scientific\\ laboratory safety\end{minipage}
& \begin{minipage}[t]{\linewidth}\raggedright MCQ + open-ended\end{minipage}
& \begin{minipage}[t]{\linewidth}\raggedright 765 MCQ;\\ 404 scenarios\end{minipage} \\
\midrule

\begin{minipage}[t]{\linewidth}\raggedright LABSHIELD\\ \cite{appendix:benchmark:labshield}\end{minipage}
& \begin{minipage}[t]{\linewidth}\raggedright OSHA standards\\ GHS\end{minipage}
& \begin{minipage}[t]{\linewidth}\raggedright laboratory safety\\ reasoning and planning\end{minipage}
& task-level VQA
& \begin{minipage}[t]{\linewidth}\raggedright 164 tasks;\\ 1,439 VQA\end{minipage} \\
\midrule

\begin{minipage}[t]{\linewidth}\raggedright UnsafeBench\\ \cite{appendix:benchmark:unsafebench}\end{minipage}
& --
& \begin{minipage}[t]{\linewidth}\raggedright image content\\ safety\end{minipage}
& \begin{minipage}[t]{\linewidth}\raggedright image-level binary\end{minipage}
& \begin{minipage}[t]{\linewidth}\raggedright 10,146 images\end{minipage} \\
\midrule

\begin{minipage}[t]{\linewidth}\raggedright SafeEditBench\\ \cite{appendix:benchmark:safeeditbench}\end{minipage}
& --
& \begin{minipage}[t]{\linewidth}\raggedright policy-adaptive\\ image safety\end{minipage}
& \begin{minipage}[t]{\linewidth}\raggedright pair-level binary\end{minipage}
& \begin{minipage}[t]{\linewidth}\raggedright 62 pairs\end{minipage} \\
\midrule

\begin{minipage}[t]{\linewidth}\raggedright ConstructionSite 10k\\ \cite{appendix:benchmark:constructionsite}\end{minipage}
& --
& \begin{minipage}[t]{\linewidth}\raggedright construction\\ safety inspection\end{minipage}
& VQA
& \begin{minipage}[t]{\linewidth}\raggedright 10,013 images\end{minipage} \\
\midrule

\begin{minipage}[t]{\linewidth}\raggedright EVADE\\ \cite{appendix:benchmark:evade}\end{minipage}
& \begin{minipage}[t]{\linewidth}\raggedright Chinese advertising law\\ platform norms\end{minipage}
& \begin{minipage}[t]{\linewidth}\raggedright e-commerce evasive\\ content detection\end{minipage}
& multi-label
& \begin{minipage}[t]{\linewidth}\raggedright 13,961 images;\\ 2,833 texts\end{minipage} \\
\midrule

\begin{minipage}[t]{\linewidth}\raggedright OS Bench\\ \cite{baselines:clue}\end{minipage}
& --
& \begin{minipage}[t]{\linewidth}\raggedright objective\\ image safety\end{minipage}
& \begin{minipage}[t]{\linewidth}\raggedright image-level binary\end{minipage}
& \begin{minipage}[t]{\linewidth}\raggedright 1,400 images\end{minipage} \\
\midrule

\begin{minipage}[t]{\linewidth}\raggedright MM-SafetyBench\\ \cite{tax:mm}\end{minipage}
& --
& \begin{minipage}[t]{\linewidth}\raggedright multimodal\\ jailbreak safety\end{minipage}
& refusal-level
& \begin{minipage}[t]{\linewidth}\raggedright 5,040 pairs\end{minipage} \\
\bottomrule
\end{tabular}
\caption{Comparison of image-based safety benchmarks. Legal or regulatory basis indicates whether the benchmark is grounded in explicit laws, regulations, or standards; domain summarizes the evaluated safety setting; output granularity reports the unit at which predictions are evaluated; and scale reports the dataset size.}
\label{tab:app:benchmark_comparison}
\end{table*}

\section{\ourbenchmark}
\label{sec:app:benchmark}

\ourbenchmark\ evaluates visual safety reasoning under concrete, domain-specific safety rules.
Table~\ref{tab:app:benchmark_comparison} compares \ourbenchmark\ with related image-based safety benchmarks in terms of legal or regulatory basis, evaluated safety domain, output granularity, and scale, highlighting its clause-level, multi-domain coverage grounded in explicit safety rules and supported by both unsafe and hard-safe cases.
\ourbenchmark\ combines real-world regulatory sources with violation cases across food safety, construction safety, platform-content safety, and advertising safety, supporting fine-grained evaluation of rule-specific visual compliance across diverse safety domains.

\subsection{Regulatory Sources}
\paragraph{FDA Food Code.}
The food-safety domain is grounded in the \textit{FDA Food Code}~\cite{law:fda_food_code_2022}, which specifies public-health safeguards for food establishments.
We use rules that can be visually assessed in inspection-like scenes, including personnel hygiene, food contamination prevention, equipment cleanliness, facility maintenance, plumbing, and refuse management.

\paragraph{NYC Building Code.}
The construction-safety domain is grounded in the \textit{NYC Building Code}~\cite{law:nyc_building_code_2022}, which defines safety requirements for buildings, construction sites, demolition sites, and means of egress.
We focus on visually observable site safety rules, especially pedestrian protection, sidewalk sheds and fences, housekeeping, concrete washout, fire protection, signage, and egress continuity.

\paragraph{Meta Community Standards.}
The platform-content safety domain is grounded in \textit{Meta Community Standards}~\cite{law:meta_community_standards}, which define policy violations for user-generated content on social platforms.
We use rules covering harmful or policy-sensitive visual content, including hateful conduct, violence and incitement, dangerous organizations, bullying and harassment, adult sexual exploitation, restricted goods, spam, and self-harm.

\paragraph{UK CAP Code.}
The advertising-content safety domain is grounded in the \textit{UK CAP Code}~\cite{law:cap_code}, which regulates non-broadcast advertising, sales promotion, and direct marketing.
We use rules for visually assessable advertising compliance, including misleading claims, required qualifications, environmental claims, financial products, alcohol, gambling, medicines and health products, electronic cigarettes, harm and offence, and marketing disclosure.

\subsection{Data Collection}
\label{sec:appendix_data_collection}
The authors collect publicly available datasets and decision repositories for each safety domain.

\paragraph{Food Safety.}
The authors collect 3,063 unique text-based cases from Chicago Food Inspections,\footnote{\url{https://data.cityofchicago.org/Health-Human-Services/Food-Inspections/4ijn-s7e5/about_data}} a City of Chicago Data Portal dataset containing inspection outcomes and reported violations for restaurants and other food establishments.
The dataset is publicly accessible through the Chicago Data Portal, whose dataset metadata refers users to the portal's terms of use for licensing.\footnote{\url{https://www.chicago.gov/city/en/narr/foia/data_disclaimer.html}}
The source is made available for public access and analysis of municipal inspection records; using these records as food-safety violation cases is consistent with that access condition.
The records describe establishments and inspection findings rather than private individual profiles, and the authors manually filter personally identifiable information before including cases in the benchmark.
The authors use the collected records in accordance with these terms and preserve source attribution.
As required by the Chicago Data Portal Terms of Use, we note that this benchmark provides cases using data that has been modified for use from its original source, www.cityofchicago.org, the official website of the City of Chicago; the City of Chicago makes no claims as to the content, accuracy, timeliness, or completeness of any of the data, which is subject to change at any time and is used at one's own risk.

\paragraph{Construction Safety.}
The authors collect 7,570 unique text-based cases from NYC DOB-ECB Violations,\footnote{\url{https://data.cityofnewyork.us/Housing-Development/DOB-ECB-Violations/6bgk-3dad}} an NYC Open Data dataset of Department of Buildings enforcement violations.
NYC Open Data provides government-produced, machine-readable datasets for public use under its open-data policies and terms.\footnote{\url{https://cityofnewyork.github.io/opendatatsm/publicpolicies.html}}
Its stated public-use context includes research, analysis, and civic applications based on government-produced data.
NYC Open Data requires agency review for confidentiality, privacy, security, and non-disclosable identifying information before publication, and the authors additionally filter personally identifiable information before including cases in the benchmark.
The authors use the collected records in accordance with these terms and preserve source attribution.

\paragraph{Platform-Content Safety.}
The platform-content domain is based on the Meta Community Standards~\cite{law:meta_community_standards} and Meta's public Community Standards Enforcement Report (CSER).\footnote{\url{https://transparency.meta.com/reports/community-standards-enforcement/}}
Because the CSER provides aggregate enforcement statistics rather than case-level user content, we use it only to guide the domain distribution.
We then synthesize text-based cases grounded in specific Community Standards rules, following the data-generation protocol of ShieldGemma~2~\cite{model:shieldgemma2} with Gemini~3.0~Flash~\cite{model:gemini_flash}.
The authors filter the generated cases for policy consistency, visual and generation quality, and use the resulting 50 curated cases for subsequent image generation.

\begin{figure}[t]
    \centering
    \includegraphics[width=\columnwidth]{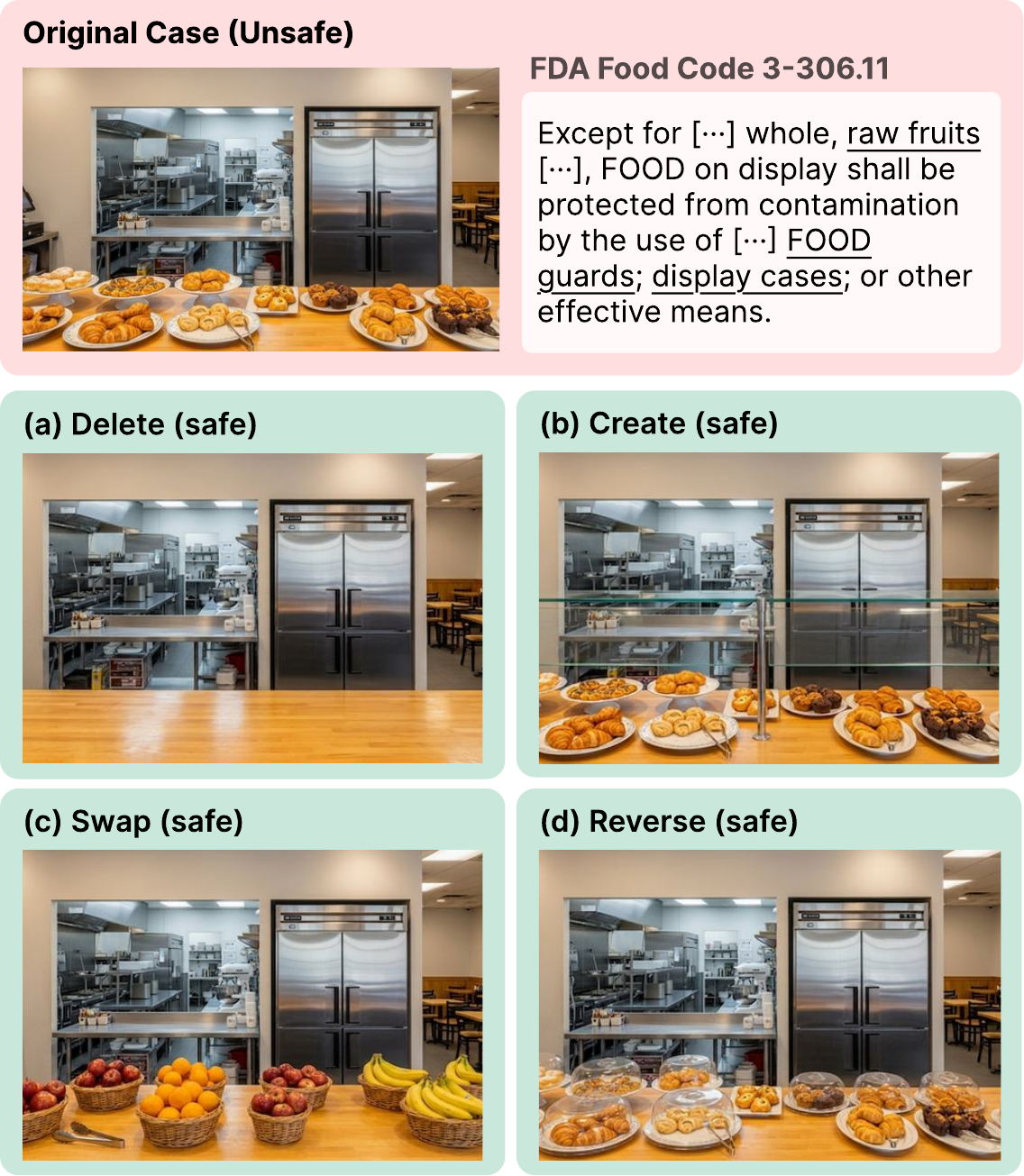}
    \caption{Hard-safe construction from violation images using four perturbations: (a) deleting a forbidden element from the scene; (b) creating a required element that is missing; (c) swapping the violating object for a confusingly similar but compliant one; (d) reversing an attribute (position, state, or orientation) of the violating element.}
    \label{fig:appendix_fig1_perturbation}
\end{figure}

\begin{figure}[t]
    \centering
    \includegraphics[width=\columnwidth]{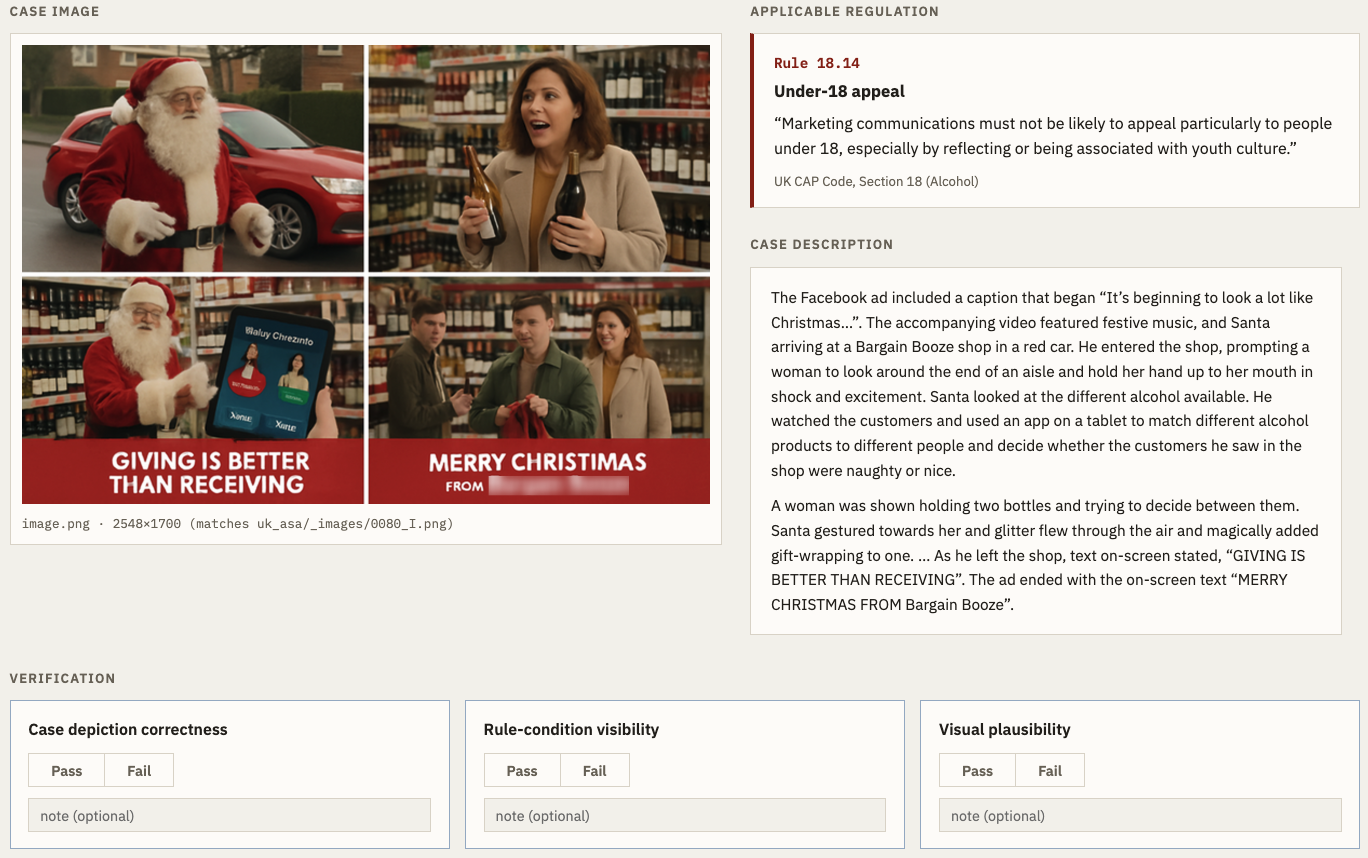}
    \caption{Verification UI used for the manual verification of \ourbenchmark.}
    \label{fig:appendix_verification_ui}
\end{figure}

\paragraph{Advertising-Content Safety.}
The authors collect 126 unique text-based cases from UK ASA rulings,\footnote{\url{https://www.asa.org.uk/}} which provide public records of how advertising rules apply after formal investigations.
The authors use these rulings in accordance with the ASA/CAP copyright statement,\footnote{\url{https://www.asa.org.uk/general/privacy-policy.html}}
which permits reproduction of website materials, unless otherwise indicated, provided that the material is reproduced accurately, not used in a misleading context, and acknowledged as ASA copyright with the title of the relevant document or publication specified.
The source is intended as a public record of advertising-rule enforcement, and using these rulings as advertising-content safety cases follows that interpretive purpose.
ASA rulings are centered on advertisers, advertising claims, and published marketing materials rather than private individual profiles, and the authors manually filter personally identifiable information before including cases in the benchmark.
The authors use the collected records in accordance with these terms and preserve source attribution.

\subsection{Image Generation and Hard-Safe Construction}

For each collected text-based case, we generate ten unsafe images using Imagen~4~\cite{model:google_imagen4} and Gemini~3 Pro Image~\cite{model:google_gemini3proimage}.
The generation prompt is grounded in the case description, while the violation label remains inherited from the original case record; for platform-content cases, the authors instead verify the label against the corresponding Meta Community Standards rule.
For hard-safe construction in the food-safety, construction-safety, and advertising-content domains, unsafe images are transformed using the perturbation procedure illustrated in Figure~\ref{fig:appendix_fig1_perturbation}: violation-critical visual evidence is removed or corrected while preserving the surrounding scene context.
%
%
This produces safe images that remain visually close to the corresponding unsafe cases, making them difficult negatives for rule-grounded safety evaluation.

\paragraph{Usage Terms, License, and Release Policy.}
The images are generated with Imagen~4~\cite{model:google_imagen4} and Gemini~3 Pro Image~\cite{model:google_gemini3proimage} under Google's generative-AI terms, which do not claim ownership of the generated content and permit research use subject to the Generative AI Prohibited Use Policy.
The images are produced through the commercial tier of the Gemini API, under whose terms Google does not use the submitted prompts or the generated outputs to train or improve its models, products, or services; the generated images carry Google's SynthID watermark and are disclosed as AI-generated.
The released images are distributed under the CC~BY~4.0 license, while the underlying source records remain governed by their respective terms of use with source attribution preserved.
For platform-content safety, we do not directly release the generated images, as they may contain platform-specific visual assets such as UI layouts, logos, and interface elements; we instead disclose the construction pipeline and specification, which can be applied to other platform policies.

\subsection{Verification}
All generated images and source cases are manually verified by the authors (Figure~\ref{fig:appendix_verification_ui}).
The verification is conducted along three axes: case depiction correctness, rule-condition visibility, and visual plausibility, with agreement rates of $97.2\%$, $96.4\%$, and $75.8\%$, respectively.
Images are filtered out when the violation or compliant state is visually ambiguous, insufficiently grounded in the source case, or degraded by low generation quality.
For the food-safety, construction-safety, and advertising-content domains, whose cases are collected from source datasets, the fine-grained violation clause label is not newly inferred during annotation but preserved verbatim from the corresponding source record; for the platform-content domain, whose cases are synthesized under a target Meta Community Standards regulation, the violation clause label is inherited from the generation condition, and the authors verify and filter cases in which the generated content does not reflect the intended label.
This keeps each unsafe label directly grounded in the underlying violation case or governing regulation.
Figures~\ref{fig:appendix_fig2_example_chicago}--\ref{fig:appendix_fig5_example_uk} show representative examples from the four source datasets, including the collected case, inherited violation label, and corresponding visual instances.

\begin{table*}[t]
\centering
\small
\resizebox{\textwidth}{!}{%
\setlength{\tabcolsep}{4pt}
\renewcommand{\arraystretch}{1.15}
\begin{tabular}{l|c>{\columncolor{gray!20}}c|c>{\columncolor{gray!20}}c|c>{\columncolor{gray!20}}c|c>{\columncolor{gray!20}}c}
\toprule
& \multicolumn{2}{c|}{\shortstack{Food\\Safety}} & \multicolumn{2}{c|}{\shortstack{Construction\\Safety}} & \multicolumn{2}{c|}{\shortstack{Platform-Content\\Safety}} & \multicolumn{2}{c}{\shortstack{Advertising-Content\\Safety}}\\
\cmidrule(lr){2-3}\cmidrule(lr){4-5}\cmidrule(lr){6-7}\cmidrule(lr){8-9}
Model & Recall & \cellcolor{white}F1-score & Recall & \cellcolor{white}F1-score & Recall & \cellcolor{white}F1-score & Recall & \cellcolor{white}F1-score \\
\midrule
Gemini 3.0 Pro & \meanstd{67.8}{0.2} & \meanstd{75.7}{0.3} & \meanstd{88.5}{2.8} & \meanstd{81.4}{2.6} & \meanstd{26.6}{1.8} & \meanstd{39.1}{4.3} & \meanstd{96.7}{3.3} & \meanstd{70.3}{2.8} \\
Gemini 3.0 Flash & \meanstd{73.6}{0.0} & \meanstd{77.4}{0.1} & \meanstd{89.8}{2.8} & \meanstd{79.5}{3.3} & \meanstd{22.8}{1.0} & \meanstd{35.2}{0.7} & \meanstd{91.9}{1.7} & \meanstd{66.8}{3.2} \\
GPT-5.4 & \meanstd{46.6}{1.2} & \meanstd{60.9}{0.8} & \meanstd{81.7}{2.7} & \meanstd{82.3}{2.5} & \meanstd{19.4}{3.9} & \meanstd{32.3}{5.6} & \meanstd{70.6}{0.5} & \meanstd{61.3}{0.8} \\
GPT-5.4 mini & \meanstd{25.9}{1.1} & \meanstd{39.7}{1.5} & \meanstd{58.5}{2.6} & \meanstd{70.7}{2.4} & \meanstd{17.7}{1.0} & \meanstd{28.9}{1.3} & \meanstd{36.8}{1.3} & \meanstd{44.6}{1.2} \\
Claude Opus 4.5 & \meanstd{62.2}{0.4} & \meanstd{69.5}{0.2} & \meanstd{90.7}{1.7} & \meanstd{78.4}{2.6} & \meanstd{23.6}{1.6} & \meanstd{35.8}{0.7} & \meanstd{85.3}{3.6} & \meanstd{63.3}{0.8} \\
Claude Sonnet 4.5 & \meanstd{38.8}{0.8} & \meanstd{49.1}{0.7} & \meanstd{83.1}{1.7} & \meanstd{76.2}{3.4} & \meanstd{25.3}{5.5} & \meanstd{39.0}{6.4} & \meanstd{73.9}{5.7} & \meanstd{59.8}{2.0} \\
Kimi K2.6 & \meanstd{52.7}{2.1} & \meanstd{64.3}{1.4} & \meanstd{90.7}{1.4} & \meanstd{83.0}{3.0} & \meanstd{17.3}{3.3} & \meanstd{27.6}{3.7} & \meanstd{82.8}{3.4} & \meanstd{63.1}{1.4} \\
Qwen 3.5 (397B) & \meanstd{49.9}{0.2} & \meanstd{58.4}{0.3} & \meanstd{90.7}{1.4} & \meanstd{78.2}{1.6} & \meanstd{21.1}{4.3} & \meanstd{34.4}{5.8} & \meanstd{69.5}{8.5} & \meanstd{60.9}{4.4} \\
\bottomrule
\end{tabular}
}
\caption{Baseline VLM performance on \ourbenchmark\ across Food Safety, Construction Safety, Platform-Content Safety, and Advertising-Content Safety over three runs.}
\label{tab:app:baseline_vlm_performance}
\end{table*}

\subsection{Dataset Statistics}
Tables~\ref{tab:app:food_category_distribution}, \ref{tab:app:building_category_distribution}, \ref{tab:app:platform_category_distribution}, and~\ref{tab:app:advertising_category_distribution} summarize the safety-category and violation-category distributions for food safety, construction safety, platform-content safety, and advertising-content safety, respectively.
The distributions show that \ourbenchmark\ covers both broad domain-level safety categories and fine-grained violation labels inherited from the real-world cases.

\subsection{Baseline VLM Performance}

\begin{figure}[t]
\centering
\includegraphics[width=\linewidth]{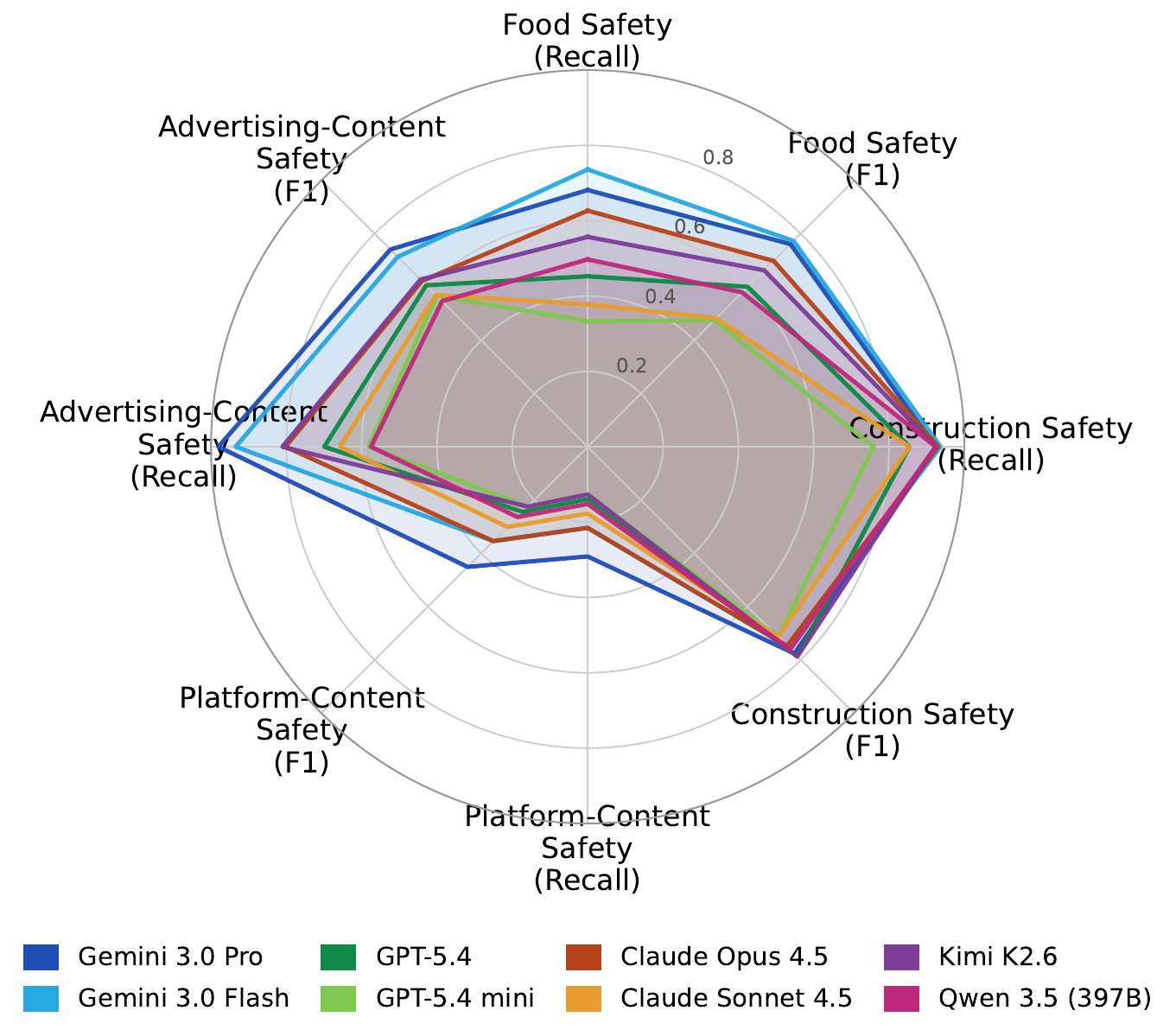}
\caption{Radar visualisation of the per-domain recall (R) and F1 numbers reported in Table~\ref{tab:app:baseline_vlm_performance} for the eight baseline VLMs. The evaluation setting is identical to the Direct Prompting baseline. Each axis is a (domain, metric) pair and the radial range is $[0, 1]$.}
\label{fig:app:baseline_vlm_performance_radar}
\end{figure}

Table~\ref{tab:app:baseline_vlm_performance} reports eight frontier VLMs evaluated on \ourbenchmark: Gemini 3.0 Pro~\cite{model:gemini_pro} and Gemini 3.0 Flash~\cite{model:gemini_flash}, GPT-5.4 and GPT-5.4 mini~\cite{model:gpt}, Claude Opus 4.5~\cite{model:claude_opus} and Claude Sonnet 4.5~\cite{model:claude_sonnet}, Kimi K2.6~\cite{model:kimi}, and Qwen 3.5 (397B)~\cite{model:qwen}.
The evaluation follows the same setting as Direct Prompting in \hyperref[sec:main_baselines]{Baselines}.
We report violation recall (R) and case-level F1 in each domain.
Figure~\ref{fig:app:baseline_vlm_performance_radar} visualises the same numbers on an eight-axis radar.

Gemini model families attain the best per-domain recall in all of four settings, while Gemini 3.0 Flash is strongest on Food F1.
The remaining frontier models show competitive performance on Construction Safety but substantially lower recall on Platform-Content Safety.

\begin{table*}[t]
\centering
\small
\setlength{\tabcolsep}{4pt}
\renewcommand{\arraystretch}{1.2}
\begin{tabular}{p{0.20\linewidth}p{0.36\linewidth}p{0.35\linewidth}}
\toprule
\textbf{Predicate Type} & \textbf{Intended Semantics} & \textbf{Grounding Evidence} \\
\midrule
\multicolumn{3}{l}{\textit{State predicates $(o,\sigma, \varnothing)$}} \\
\midrule
Visual state
& $o$ exhibits the visual state or attribute $s$.
& Detected object bounding box and VLM state module. \\

Textual semantic state
& $o$ contains text whose meaning or referent matches $s$.
& OCR output and VLM state module. \\
\midrule
\multicolumn{3}{l}{\textit{Relation predicates $(o_1,\rho,o_2)$}} \\
\midrule
\texttt{any} / \texttt{None}
& No relational constraint.
& -- \\

\texttt{inside}
& $o_1$ is spatially contained within $o_2$.
& Box containment, region overlap, and depth consistency. \\

\texttt{near}
& $o_1$ is physically close to $o_2$.
& Image-plane distance and depth-aware proximity. \\

\texttt{adjacent}
& $o_1$ is next to or directly touching $o_2$.
& Boundary distance, overlap, and depth consistency. \\

\texttt{above}
& $o_1$ is vertically above $o_2$ in the scene.
& Vertical ordering, overlap suppression, and depth consistency. \\

\texttt{below}
& $o_1$ is vertically below $o_2$ in the scene.
& Vertical ordering, overlap suppression, and depth consistency. \\

\texttt{on}
& $o_1$ rests on or is supported by $o_2$.
& Vertical contact, horizontal support overlap, and depth consistency. \\

\texttt{at\_edge\_of}
& $o_1$ lies at the boundary region of $o_2$.
& Boundary overlap, relative position, and depth consistency. \\

\texttt{covering}
& $o_1$ occludes or covers a substantial part of $o_2$.
& Box overlap, containment ratio, and foreground--background depth ordering. \\
\bottomrule
\end{tabular}
\caption{Specification of pre-defined grounding predicates used for scene-graph node and edge labeling. State predicates assign object-local labels $(o,\sigma,\varnothing)$ using visual or textual evidence, while relation predicates assign edge labels $(o_1,\rho,o_2)$ from localized geometric and depth evidence. Concrete thresholds and module-level configurations are treated as implementation parameters, while the semantic role of each predicate is fixed across domains.}
\label{tab:appendix_predefined_grounding_specs}
\end{table*}

\begin{table*}[t]
\centering
\small
\setlength{\tabcolsep}{4pt}
\renewcommand{\arraystretch}{1.15}
\begin{tabular}{l|c>{\columncolor{gray!20}}c|c>{\columncolor{gray!20}}c|c>{\columncolor{gray!20}}c|c>{\columncolor{gray!20}}c}
\toprule
& \multicolumn{2}{c|}{\shortstack{Food\\Safety}} & \multicolumn{2}{c|}{\shortstack{Construction\\Safety}} & \multicolumn{2}{c|}{\shortstack{Platform-Content\\Safety}} & \multicolumn{2}{c}{\shortstack{Advertising-Content\\Safety}} \\
\cmidrule(lr){2-3} \cmidrule(lr){4-5} \cmidrule(lr){6-7} \cmidrule(lr){8-9}
Method & Low & \cellcolor{white}High & Low & \cellcolor{white}High & Low & \cellcolor{white}High & Low & \cellcolor{white}High \\
\midrule
Direct Prompting   & \meanstd{46.9}{3.9} & \meanstd{46.0}{2.4} & \meanstd{78.5}{3.7} & \meanstd{67.7}{3.0} & \meanstd{53.9}{3.9} & \meanstd{6.5}{5.3} & \meanstd{51.3}{0.4} & \meanstd{67.1}{0.6} \\
ViperGPT           & \meanstd{79.1}{0.8} & \meanstd{66.7}{0.4} & \meanstd{75.4}{0.3} & \meanstd{68.4}{0.0} & \meanstd{51.3}{2.7} & \meanstd{49.1}{4.2} & \meanstd{47.6}{0.4} & \meanstd{46.2}{1.3} \\
ETA                & \meanstd{88.2}{0.1} & \meanstd{73.3}{0.9} & \meanstd{85.8}{0.2} & \meanstd{71.3}{1.7} & \meanstd{71.2}{2.7} & \meanstd{43.6}{8.2} & \meanstd{56.9}{1.7} & \meanstd{56.9}{2.9} \\
SafeCLIP           & \meanstd{87.6}{0.7} & \meanstd{72.0}{0.6} & \meanstd{92.1}{1.4} & \meanstd{49.3}{2.1} & \meanstd{66.3}{4.3} & \meanstd{36.2}{5.1} & \meanstd{49.3}{2.0} & \meanstd{57.9}{2.5} \\
CLUE               & \meanstd{81.6}{0.6} & \meanstd{59.1}{0.7} & \meanstd{70.7}{0.8} & \meanstd{68.6}{0.7} & \meanstd{57.0}{1.8} & \meanstd{51.6}{1.2} & \meanstd{78.5}{1.9} & \meanstd{27.1}{0.6} \\
CompAgent          & \meanstd{83.9}{1.1} & \meanstd{69.9}{2.1} & \meanstd{81.9}{1.1} & \meanstd{67.7}{1.7} & \meanstd{76.5}{3.6} & \meanstd{38.8}{6.0} & \meanstd{56.7}{3.1} & \meanstd{53.0}{1.0} \\
\midrule
\ourmodel          & \meanstd{89.0}{1.9} & \meanstd{79.0}{1.6} & \meanstd{84.7}{1.5} & \meanstd{83.5}{1.0} & \meanstd{79.1}{1.3} & \meanstd{68.1}{6.7} & \meanstd{81.5}{2.8} & \meanstd{75.8}{0.9} \\
\bottomrule
\end{tabular}
\caption{F1-score (mean over three runs) on \ourbenchmark\ for the \textsc{Low} and \textsc{High} complexity tertiles of the per-domain score $C_d(r)$ (Eq.~(\ref{eq:complexity})), evaluated with Gemma~4~26B.}
\label{tab:appendix_complexity_f1}
\end{table*}

\section{Implementation Details}
All pipelines, including the baselines and \ourmodel, are composed with LangChain over an OpenAI-compatible interface. The visual modules use HuggingFace Transformers (with PyTorch) for open-vocabulary detection (SAM~3~\cite{model:sam3}), image--text scoring (SigLIP~2~\cite{model:siglip2}), and monocular depth (Depth Anything~3~\cite{model:depth3}), and Google Document AI~\cite{model:ocr} for OCR.

\subsection{Baselines}
\subsubsection{ViperGPT}
ViperGPT~\cite{baselines:vipergpt} is included as an executable visual-reasoning baseline, testing whether code generation over vision APIs is sufficient for safety-rule compliance.
We follow the official implementation and adapt its program-generation interface to our image--rule setting, where the model generates \texttt{execute\_command(images, regulations, api)} for a structured violation verdict.
The provided \texttt{ImagePatch}-style API wraps the same visual foundation models used in our main experiments (SAM~3, SigLIP~2, Depth Anything~3, and Google's Document OCR) and additionally allows VLM calls for open-ended or yes/no visual checks.
%

\subsubsection{ETA}
ETA~\cite{baseline:eta} is included as an inference-time safety-alignment baseline that first evaluates a VLM judgment and invokes alignment only when the judgment is uncertain.
We follow the official implementation and adapt its evaluate-then-align pipeline to our image--rule setting: an applicability evaluator filters clearly irrelevant pairs, a VLM produces a structured violation judgment, and uncertain cases are resolved by selecting the best grounded candidate from a sampled pool.
We use the paper-default best-of-\(N=5\), with a SigLIP applicability threshold of \(0.05\), a reward-gate band of \([0.30, 0.70]\), and candidate-ranking weights \(0.4/0.5/0.1\) for grounding, soundness, and SAM~3 detection, respectively.
\subsubsection{SafeCLIP}
SafeCLIP~\cite{rel:baseline:zero} is included as a representation-alignment baseline, testing whether image--policy similarity can serve as a zero-shot safety signal before explicit rule reasoning.
We construct a safety concept bank from the target regulation, add a neutral category, embed the image and text descriptors with SigLIP~2, and flag the image when the highest non-neutral regulation score exceeds both the toxicity threshold and the neutral score.
When flagged, the paper's generic safety template is prepended to the original VLM query before producing the final structured violation judgment.
We use \(K=5\) descriptors per category, softmax temperature scale \(\sigma=100\), toxicity threshold \(\tau=0.6\), VLM temperature \(0.3\), and aggregate multi-image cases by the maximum regulation score across images.
\subsubsection{CLUE}
CLUE~\cite{baselines:clue} is included as an objectified-precondition baseline, testing whether safety rules can be converted into visually checkable conditions before judgment.
Our CLUE implementation follows the objectification and centric-region masking pipeline, while using VLM-based Yes/No token probabilities and reasoning fallback for precondition judgment.
We first objectify each regulation and decompose it into preconditions; this offline decomposition and optimization stage is performed with GPT-5.4.
At inference time, centric phrases for the preconditions are grounded with SAM~3, and each precondition is judged using VLM Yes/No token probabilities over the original image, the no-image prior, and the centric-region-masked image.
%
\subsubsection{CompAgent}
CompAgent~\cite{baselines:compagent} is included as a policy-guided agentic tool-use baseline, testing whether iterative evidence collection is sufficient for visual compliance verification.
We adapt its two-agent workflow: a text-only Planning Agent selects tools in a ReAct loop, and a multimodal CVAgent produces the final structured verdict from the image, regulation, and accumulated evidence.
The main agent components use the same inference VLM as the other baselines, while the tool suite includes VLM-based summarization, SAM~3/SigLIP~2 object detection, Google's Document OCR, SigLIP~2 scoring as a Safe-CLIP substitute, and LlamaGuard~\cite{model:llamaguard4} as a callable evidence tool.
We use at most 10 ReAct~\cite{method:one:react} steps with decoding temperature \(0.0\).
\subsection{\ourmodel}
\label{sec:appendix_ourmodel_implementation}

\subsubsection{\stepone}
\label{sec:appendix_stepone_details}

\stepone\ follows the iterative decomposition and verification procedure in
Section~\ref{sec:method:stepone}. At each sub-rule $r$ (also at $R$), decomposer $\mathbb{D}$ proposes
$(\varphi_r^{(t)}, \{\hat r_1^{(t)}, \ldots, \hat r_k^{(t)}\})$ and verifier $\mathbb{V}$ returns
$(s_r^{(t)}, \eta_r^{(t)}, a_r^{(t)})$. Refinement terminates when
$a_r^{(t)} = \mathtt{True}$, $s_r^{(t)}$ saturates, or $s_r^{(t)}$ reaches $s_\mathrm{max}$; the default patience is set to
three iterations. Figures~\ref{fig:appendix_decomposer_prompt}
and~\ref{fig:appendix_verifier_prompt} provide the
decomposer and verifier prompts.

\paragraph{Pre-defined Grounding.}
\label{sec:appendix_predefined_grounding_rules}

The Selection agent $\mathbb{V}_{\mathrm{sel}}$ produces the requested objects
$O_{\mathrm{req}}$ and relations $\Gamma_{\mathrm{req}}$ for each atomic
proposition from the regulation and the image, following the prompt in
Figure~\ref{fig:appendix_selection_prompt}.
Given $O_{\mathrm{req}}$ and $\Gamma_{\mathrm{req}}$,
$\mathrm{Label}_o$ assigns state labels $s_o$ and $\mathrm{Label}_\rho$ assigns
relation labels $v$ for constructing $\mathcal G^{(I)}=(\mathcal V^{(I)},\mathcal E^{(I)})$.
Table~\ref{tab:appendix_predefined_grounding_specs} summarizes the predicate
specification.

Satisfied relation predicates instantiate scene-graph edges for subsequent
rule-tree evaluation.

\subsubsection{Tool-Aware Visual Grounding}
\label{sec:appendix_tool_aware_grounding_details}

Tool-Aware Visual Grounding follows Section~\ref{sec:method:tool_aware}. When
$c_o<\tau$, $\mathrm{Align}(o;I,\mathcal{C})$ invokes
$\mathbb{V}_{\mathrm{ground}}$ to obtain an accepted substitute $k_o^{\star}$,
which is cached as $\mathcal{C}[o]$. Figure~\ref{fig:appendix_grounding_prompt}
provides the $\mathbb{V}_{\mathrm{ground}}$ prompt. This behavior is evaluated in
Section~\ref{sec:appendix_fallback_accumulation}.

\section{Experimental Details}
\label{sec:appendix_experimental_details}

\begin{table*}[t]
\centering
\small
\resizebox{\textwidth}{!}{%
\setlength{\tabcolsep}{4pt}
\renewcommand{\arraystretch}{1.15}
\begin{tabular}{l|l|c>{\columncolor{gray!20}}c|c>{\columncolor{gray!20}}c|c>{\columncolor{gray!20}}c|c>{\columncolor{gray!20}}c}
\toprule
& & \multicolumn{2}{c|}{\shortstack{Food\\Safety}} & \multicolumn{2}{c|}{\shortstack{Construction\\Safety}} & \multicolumn{2}{c|}{\shortstack{Platform-Content\\Safety}} & \multicolumn{2}{c}{\shortstack{Advertising-Content\\Safety}}\\
\cmidrule(lr){3-4}\cmidrule(lr){5-6}\cmidrule(lr){7-8}\cmidrule(lr){9-10}
Method & Inference Model & Recall & \cellcolor{white}F1-score & Recall & \cellcolor{white}F1-score & Recall & \cellcolor{white}F1-score & Recall & \cellcolor{white}F1-score \\
\midrule
\multirow{2}{*}{Direct Prompting}
 & Gemma~4~26B             & \meanstd{34.5}{2.4} & \meanstd{45.4}{2.3} & \meanstd{72.0}{1.6} & \meanstd{78.6}{1.5} & \meanstd{16.5}{3.6} & \meanstd{26.6}{4.4} & \meanstd{57.5}{1.3} & \meanstd{57.1}{0.8} \\
 & GPT-5.4-mini         & \meanstd{25.9}{1.1} & \meanstd{39.7}{1.5} & \meanstd{58.5}{2.6} & \meanstd{70.7}{2.4} & \meanstd{17.7}{1.0} & \meanstd{28.9}{1.3} & \meanstd{36.8}{1.3} & \meanstd{44.6}{1.2} \\
\midrule
\multirow{2}{*}{ViperGPT}
 & Gemma~4~26B             & \meanstd{73.1}{0.4} & \meanstd{72.5}{0.4} & \meanstd{84.0}{0.0} & \meanstd{74.0}{0.3} & \meanstd{43.9}{1.6} & \meanstd{53.3}{4.0} & \meanstd{60.6}{0.3} & \meanstd{45.5}{0.7} \\
 & GPT-5.4-mini         & \meanstd{79.9}{1.5} & \meanstd{72.5}{0.9} & \meanstd{94.1}{0.0} & \meanstd{74.9}{0.2} & \meanstd{78.1}{0.6} & \meanstd{66.1}{0.4} & \meanstd{87.4}{1.6} & \meanstd{50.3}{0.4} \\
\midrule
\multirow{2}{*}{ETA}
 & Gemma~4~26B             & \meanstd{83.7}{0.8} & \meanstd{82.6}{0.2} & \meanstd{95.8}{1.2} & \meanstd{83.6}{1.2} & \meanstd{39.7}{4.3} & \meanstd{52.0}{7.5} & \meanstd{73.4}{0.3} & \meanstd{57.4}{1.2} \\
 & GPT-5.4-mini         & \meanstd{62.5}{0.8} & \meanstd{72.1}{0.9} & \meanstd{69.5}{0.8} & \meanstd{77.5}{1.2} & \meanstd{51.5}{2.4} & \meanstd{61.3}{1.3} & \meanstd{54.5}{3.0} & \meanstd{49.7}{2.4} \\
\midrule
\multirow{2}{*}{SafeCLIP}
 & Gemma~4~26B             & \meanstd{82.9}{1.2} & \meanstd{81.6}{0.2} & \meanstd{79.0}{0.0} & \meanstd{84.2}{0.7} & \meanstd{41.8}{2.1} & \meanstd{53.5}{2.1} & \meanstd{71.9}{0.9} & \meanstd{55.2}{1.7} \\
 & GPT-5.4-mini         & \meanstd{60.9}{0.2} & \meanstd{71.5}{0.6} & \meanstd{66.1}{0.8} & \meanstd{76.4}{0.5} & \meanstd{69.6}{2.1} & \meanstd{73.8}{1.6} & \meanstd{57.1}{1.3} & \meanstd{52.1}{1.2} \\
\midrule
\multirow{2}{*}{CLUE}
 & Gemma~4~26B             & \meanstd{66.0}{0.8} & \meanstd{71.7}{0.6} & \meanstd{87.1}{1.0} & \meanstd{72.1}{0.7} & \meanstd{53.2}{2.5} & \meanstd{57.9}{2.6} & \meanstd{55.3}{1.2} & \meanstd{44.2}{0.3} \\
 & GPT-5.4-mini         & \meanstd{58.7}{1.0} & \meanstd{64.3}{1.2} & \meanstd{88.5}{1.9} & \meanstd{70.9}{1.0} & \meanstd{66.2}{2.9} & \meanstd{63.5}{1.7} & \meanstd{54.0}{2.0} & \meanstd{38.9}{1.3} \\
\midrule
\multirow{2}{*}{CompAgent}
 & Gemma~4~26B             & \meanstd{76.8}{1.9} & \meanstd{77.9}{1.0} & \meanstd{91.6}{1.8} & \meanstd{80.6}{1.2} & \meanstd{54.9}{3.2} & \meanstd{59.5}{2.1} & \meanstd{64.3}{2.4} & \meanstd{53.7}{1.4} \\
 & GPT-5.4-mini         & \meanstd{64.2}{2.0} & \meanstd{71.4}{2.3} & \meanstd{76.5}{1.8} & \meanstd{82.3}{1.2} & \meanstd{52.3}{2.2} & \meanstd{56.3}{1.6} & \meanstd{79.3}{4.8} & \meanstd{57.7}{1.3} \\
\midrule
\multirow{2}{*}{\ourmodel}
 & Gemma~4~26B    & \meanstd{89.3}{1.2} & \meanstd{\textbf{84.4}}{3.5} & \meanstd{92.9}{0.8} & \meanstd{\textbf{86.4}}{0.1} & \meanstd{70.0}{4.0} & \meanstd{\textbf{69.0}}{2.6} & \meanstd{69.8}{0.5} & \meanstd{\textbf{74.9}}{3.6} \\
 & GPT-5.4-mini   & \meanstd{88.3}{1.4} & \meanstd{\textbf{84.5}}{0.6} & \meanstd{95.2}{0.4} & \meanstd{\textbf{88.2}}{0.4} & \meanstd{77.9}{3.4} & \meanstd{\textbf{74.6}}{2.4} & \meanstd{80.0}{1.5} & \meanstd{\textbf{79.8}}{1.6} \\\bottomrule
\end{tabular}
}
\caption{Effect of the inference VLM on baseline performance. Each method is evaluated with Gemma~4~26B~\cite{model:gemma4} and GPT-5.4-mini~\cite{model:gpt} across the four safety domains of \ourbenchmark.}
\label{tab:appendix_inference_generalizability}
\end{table*}
\begin{table*}[t]
\centering
\small
\setlength{\tabcolsep}{8pt}
\renewcommand{\arraystretch}{1.3}
\begin{tabular}{lcccc}
\toprule
\textbf{Removed Component}
 & \textbf{\shortstack{Food\\Safety}}
 & \textbf{\shortstack{Construction\\Safety}}
 & \textbf{\shortstack{Platform-Content\\Safety}}
 & \textbf{\shortstack{Advertising-Content\\Safety}} \\
\midrule
\ourmodel\ (Full)        & \meanstd{84.4}{3.5} & \meanstd{86.4}{0.1} & \meanstd{69.0}{2.6} & \meanstd{74.9}{3.6} \\
\midrule
Scene-Grounded Execution & \meanstd{75.9}{1.4} & \meanstd{80.5}{4.4} & \meanstd{64.7}{4.6} & \meanstd{53.4}{9.6} \\
Verifier Loop            & \meanstd{74.6}{3.1} & \meanstd{62.2}{15.7} & \meanstd{46.3}{1.8} & \meanstd{45.1}{12.8} \\
Tool-Aware Grounding     & \meanstd{76.2}{4.6} & \meanstd{75.5}{6.8} & \meanstd{64.6}{2.9} & \meanstd{46.5}{12.8} \\
Relation Grounding       & \meanstd{77.7}{3.6} & \meanstd{52.0}{7.6} & \meanstd{45.5}{2.8} & \meanstd{44.9}{13.4} \\
\bottomrule
\end{tabular}
\caption{Ablation study of \ourmodel\ on \ourbenchmark, reported as mean F1-score over three runs per domain. Each row below the first reports the score when the listed component is removed from the full pipeline. All variants are evaluated with Gemma~4~26B~\cite{model:gemma4} as the underlying inference VLM.}
\label{tab:appendix_ablation_study}
\end{table*}

\begin{table}[t]
\centering
\small
\setlength{\tabcolsep}{4pt}
\renewcommand{\arraystretch}{1.15}
\begin{tabular*}{\columnwidth}{@{\extracolsep{\fill}}l|l c c c@{}}
\toprule
Visual cue & Rule & DP & Ours & $\Delta$ \\
\midrule
\multirow{6}{*}{Hate symbols}
 & S1 Coord. Harm & 25.0 & 0.0  & 100\%$\downarrow$  \\
 & S2 Dang. Orgs. & 91.7 & 29.2 & 68.2\%$\downarrow$ \\
 & S5 Violence    & 44.0 & 4.0  & 90.9\%$\downarrow$ \\
 & S7 Bullying    & 56.2 & 0.0  & 100\%$\downarrow$  \\
 & S13 Hateful    & 78.9 & 5.3  & 93.3\%$\downarrow$ \\
 & S21 Misinfo.   & 28.1 & 0.0  & 100\%$\downarrow$  \\
\midrule
\multirow{6}{*}{Blood / Injury}
 & S1 Coord. Harm & 42.1 & 0.0  & 100\%$\downarrow$  \\
 & S2 Dang. Orgs. & 35.0 & 10.0 & 71.4\%$\downarrow$ \\
 & S5 Violence    & 10.0 & 0.0  & 100\%$\downarrow$  \\
 & S7 Bullying    & 50.0 & 0.0  & 100\%$\downarrow$  \\
 & S13 Hateful    & 54.5 & 0.0  & 100\%$\downarrow$  \\
 & S21 Misinfo.   & 30.0 & 0.0  & 100\%$\downarrow$  \\
\midrule
\multirow{6}{*}{Protest / Riot}
 & S1 Coord. Harm & 57.3 & 20.0 & 65.1\%$\downarrow$ \\
 & S2 Dang. Orgs. & 69.7 & 5.3  & 92.5\%$\downarrow$ \\
 & S5 Violence    & 36.0 & 24.0 & 33.3\%$\downarrow$ \\
 & S7 Bullying    & 36.8 & 0.0  & 100\%$\downarrow$  \\
 & S13 Hateful    & 30.8 & 0.0  & 100\%$\downarrow$  \\
 & S21 Misinfo.   & 53.3 & 2.7  & 95.0\%$\downarrow$ \\
\midrule
\multirow{6}{*}{Targeted text}
 & S1 Coord. Harm & 28.6 & 4.3  & 84.8\%$\downarrow$ \\
 & S2 Dang. Orgs. & 60.3 & 8.4  & 86.0\%$\downarrow$ \\
 & S5 Violence    & 33.9 & 7.4  & 78.0\%$\downarrow$ \\
 & S7 Bullying    & 39.8 & 0.0  & 100\%$\downarrow$  \\
 & S13 Hateful    & 45.3 & 3.4  & 92.5\%$\downarrow$ \\
 & S21 Misinfo.   & 27.9 & 1.1  & 96.1\%$\downarrow$ \\
\midrule
\multirow{5}{*}{Property damage}
 & S1 Coord. Harm & 77.4 & 38.7 & 50.0\%$\downarrow$ \\
 & S2 Dang. Orgs. & 61.3 & 9.7  & 84.2\%$\downarrow$ \\
 & S5 Violence    & 21.4 & 21.4 & 0.0\%             \\
 & S7 Bullying    & 20.0 & 0.0  & 100\%$\downarrow$  \\
 & S21 Misinfo.   & 51.6 & 3.2  & 93.8\%$\downarrow$ \\
\midrule
\multirow{5}{*}{Child}
 & S1 Coord. Harm & 0.0  & 6.2  & --                \\
 & S2 Dang. Orgs. & 56.2 & 6.2  & 88.9\%$\downarrow$ \\
 & S5 Violence    & 12.5 & 0.0  & 100\%$\downarrow$  \\
 & S13 Hateful    & 83.3 & 0.0  & 100\%$\downarrow$  \\
 & S21 Misinfo.   & 12.5 & 0.0  & 100\%$\downarrow$  \\
\midrule
\multicolumn{2}{l}{\textbf{Overall}} & \textbf{15.5} & \textbf{3.2} & \textbf{80.4\%}$\boldsymbol{\downarrow}$ \\
\bottomrule
\end{tabular*}
\caption{Visual-sensitivity-induced bias per rule.}
\label{tab:appendix_visual_sensitivity_bias}
\end{table}

\begin{table}[t]
\centering
\small
\setlength{\tabcolsep}{3pt}
\renewcommand{\arraystretch}{1.15}
\begin{tabular*}{\columnwidth}{@{\extracolsep{\fill}}l|ccc@{}}
\toprule
Rule & DP & Ours & $\Delta$ \\
\midrule
S2 Dang. Orgs.       & 52.8 & 7.8 & 85.2\%$\downarrow$  \\
S13 Hateful          & 40.1 & 3.0 & 92.5\%$\downarrow$  \\
S7 Bullying          & 33.9 & 0.0 & 100\%$\downarrow$   \\
S5 Violence          & 27.7 & 6.1 & 78.0\%$\downarrow$  \\
S1 Coord. Harm       & 26.4 & 5.2 & 80.4\%$\downarrow$  \\
S21 Misinfo.         & 23.9 & 0.9 & 96.1\%$\downarrow$  \\
S6 Adult Exploit.    & 3.3  & 0.2 & 92.9\%$\downarrow$  \\
S15 Graphic Content  & 2.8  & 8.2 & 191.7\%$\uparrow$   \\
S22 Spam             & 2.1  & 1.9 & 11.1\%$\downarrow$  \\
S4 Restricted Goods  & 1.9  & 0.0 & 100\%$\downarrow$   \\
S10 Self-Injury      & 1.2  & 0.2 & 80.0\%$\downarrow$  \\
S8 Child Exploit.    & 0.2  & 0.0 & 100\%$\downarrow$   \\
S9 Human Exploit.    & 0.2  & 0.0 & 100\%$\downarrow$   \\
\midrule
\textbf{Overall}     & \textbf{12.1} & \textbf{2.0} & \textbf{83.8\%}$\boldsymbol{\downarrow}$ \\
\bottomrule
\end{tabular*}
\caption{Rule-induced bias per rule.}
\label{tab:appendix_rule_induced_bias}
\end{table}

\subsection{Robustness to Rule Complexity}
\label{sec:appendix_robustness_details}

This subsection details the protocol underlying the robustness-to-complexity analysis in Section~\ref{sec:analysis_robustness}.

\paragraph{Per-Rule Complexity Score.}
For each rule $r$, we extract seven non-negative features from its regulation clause:
\begin{itemize}\setlength\itemsep{0.1em}
\item $f_{\mathrm{cond}}$: conditionals (\emph{if, when, unless...}).
\item $f_{\mathrm{enum}}$: enumerated lists.
\item $f_{\mathrm{xref}}$: cross-references to other rules.
\item $f_{\mathrm{neg}}$: negations (\emph{not, no, never...}).
\item $f_{\mathrm{ao}}$: connectives \emph{and}/\emph{or}.
\item $f_{\mathrm{conc}}$: concrete objects (\emph{door, sign, wall...}).
\item $f_{\mathrm{exc}}$: exceptions (\emph{except, notwithstanding...}).
\end{itemize}
Each captures a distinct facet of logical or structural complexity. We stack them into a feature vector
\begin{equation}
\mathbf{f}\!=\!(f_{\mathrm{cond}},\! f_{\mathrm{enum}},\! f_{\mathrm{xref}},\! f_{\mathrm{neg}},\! f_{\mathrm{ao}},\! f_{\mathrm{conc}},\! f_{\mathrm{exc}})
\label{eq:feature}
\end{equation}
and define the domain-specific complexity score as their weighted sum:
\begin{equation}
C(r) = {\mathbf{w}^{\star}}^{\top} \mathbf{f}(r),
\label{eq:complexity}
\end{equation}
where the non-negative weights are fit separately for each domain to approximate the rule-level F1 degradation averaged over all evaluated methods:
\begin{equation}
\mathbf{w}^{\star}
=
\underset{\mathbf{w}\ge 0}{\arg\min}
\sum_{r \in \mathcal R}
\left(y_d(r)-\mathbf{w}^{\top}\mathbf{f}(r)\right)^2.
\label{eq:complexity_fit}
\end{equation}
Here $y_d(r)$ is the observed rule-level F1 drop averaged over the six baselines of Section~\ref{sec:main_baselines} and \ourmodel, and
\begin{equation}
    \mathcal R = \{R\} \cup \operatorname{sub}^\star(R)
\end{equation}
where
\begin{equation}
    \operatorname{sub}^\star(R) = \bigcup_{\hat r \in \operatorname{sub}(R)}\{ \{\hat r\} \cup \operatorname{sub}^\star(\hat r) \}.
\end{equation}
Within each domain, the rules are ranked by $C(r)$ and split into three bins: \textsc{Low}, \textsc{Medium}, and \textsc{High}. Per-bin F1 scores across the four domains are reported in Table~\ref{tab:appendix_complexity_f1}.

\paragraph{Robustness Index.}
We define the Robustness Index as a scalar summary of a method's F1 trajectory across complexity bins:
\begin{equation}
F_1^{\textsc{High}} \cdot \left(1 - \operatorname{ReLU}\left(\frac{F_1^{\textsc{Low}} - F_1^{\textsc{High}}}{F_1^{\textsc{Low}}}\right)\right).
\label{eq:ri}
\end{equation}
The first factor rewards absolute performance on the hardest rules; the second penalizes the relative drop from the easiest bin and is clamped at $F_1^{\textsc{High}}$.

\subsection{Visual Sensitivity Bias}
\label{sec:appendix_visual_sensitivity_details}

This subsection details the protocol underlying the visual-sensitivity-bias analysis in Section~\ref{sec:analysis_biases}.

\paragraph{Image Set and Visual Cue Categorisation.}
We use all 434 violation images from the platform-content domain, spanning 46 distinct base scenes. Each image is classified by a separate Gemini~3.0 Pro~\cite{model:gemini_pro} judge into one of seven \emph{visual cue categories}---hate symbols, weapons, blood/injury, protest/riot, targeted text, property damage, child---or the residual class \emph{(no risk cue)} when no such cue is visible. An image may be assigned multiple categories when several cues co-occur; the residual class is mutually exclusive with the others.

\paragraph{(Image, Rule) Querying.}
We query the end-to-end Direct Prompting baseline and \ourmodel\ on image--rule pairs formed by pairing each sampled image with each of the 16 top-level Meta Community Standards rules. The spurious violation rate is computed over pairs whose rule is not annotated as applicable to the image.

\paragraph{Per-Rule Decomposition.}
Table~\ref{tab:appendix_visual_sensitivity_bias} reports the full category--rule matrix together with the per-cell relative change $\Delta$; the abbreviated rule labels are expanded in the caption. Cells in which neither method records any spurious violation in a given category are omitted.

\subsection{Rule-Induced Bias}
\label{sec:appendix_rule_induced_details}

This subsection details the protocol underlying the rule-induced-bias analysis in Section~\ref{sec:analysis_biases} and reports the full per-regulation breakdown summarised in Table~\ref{tab:main_rule_induced_bias}.

\begin{figure*}[t]
    \centering
    \includegraphics[width=\textwidth]{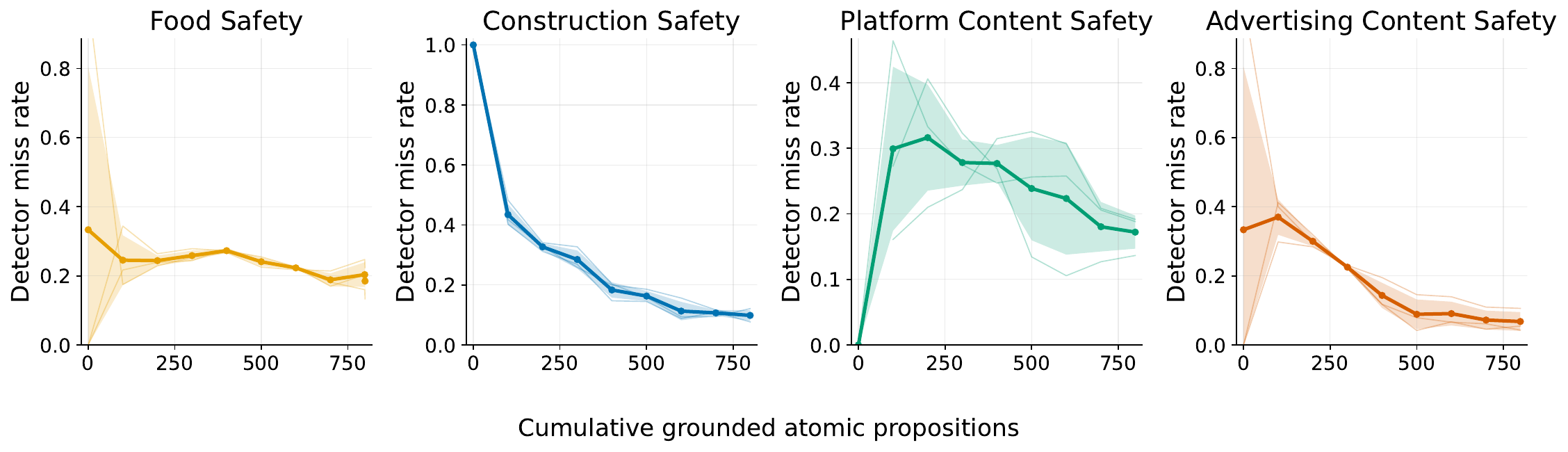}
    \caption{
     The y-axis shows the mean miss rate (confidence $< \tau$) over three independent runs as the per-domain substitute-keyword cache $\mathcal{C}$ accumulates, computed over $p_i$ whose target is visibly present in the image.
    }
    \label{fig:appendix_continual_grounding}
\end{figure*}

\paragraph{(Image, Rule) Querying.}
We use the same image--rule querying protocol as in the visual-sensitivity-bias analysis.

\paragraph{Per-Regulation Breakdown.}
Table~\ref{tab:appendix_rule_induced_bias} reports all 16 regulations ranked by DP spurious violation rate. Under DP, six regulations spuriously flag more than $20\%$ of unrelated images, peaking at $52.8\%$ for Dangerous Organizations and Individuals. \ourmodel\ removes most of these spurious violations by replacing each monolithic regulation with value-neutral atomic checks, lowering the overall rate from $12.1\%$ to $2.0\%$. The single exception is Violent and Graphic Content (S15), where decomposition slightly increases the rate on an already-low base. 

\subsection{Ablation Study}
\label{sec:appendix_ablation_study}

Table~\ref{tab:appendix_ablation_study} reports the per-domain F1-score of each \ourmodel\ variant on \ourbenchmark\ using Gemma~4~26B as the underlying inference VLM. Each row removes a single design component while keeping the remaining pipeline intact.

\paragraph{w/o \steptwo.} The compiled tree is preserved, but each atomic proposition is judged by the VLM over the raw image rather than the scene graph $\mathcal G^{(I)}$.
\paragraph{w/o Verifier Loop.} The decomposer $\mathbb{D}$ is run once per node without verifier feedback $\eta_n^{(t)}$ or score-driven re-sampling.
\paragraph{w/o Tool-Aware Grounding.} The detector is used as-is without the abstraction-gap and lexical-gap alignment described in Section~\ref{sec:method:tool_aware}.
\paragraph{w/o Relation Grounding.} Only object-state propositions $(o,\sigma,\varnothing)$ are evaluated; relational propositions $(o_1,\rho,o_2)$ are discarded.

\subsection{Sensitivity to Grounding Confidence}
\label{sec:appendix_grounding_confidence}

\begin{table}[t]
\centering
\small
\setlength{\tabcolsep}{4pt}
\renewcommand{\arraystretch}{1.2}
\resizebox{\columnwidth}{!}{%
\begin{tabular}{c|cccc}
\toprule
$\tau$ & Food & Constr. & Platform & Advert. \\
\midrule
0.4 & \meanstd{82.3}{2.4} & \meanstd{82.2}{2.5} & \meanstd{66.1}{1.6} & \meanstd{60.9}{3.2} \\
0.5 & \meanstd{83.4}{1.7} & \meanstd{82.1}{2.3} & \meanstd{65.8}{2.3} & \meanstd{65.0}{2.8} \\
0.6 & \meanstd{83.4}{2.0} & \meanstd{85.0}{2.5} & \meanstd{68.9}{1.5} & \meanstd{68.1}{2.6} \\
0.7 & \meanstd{85.6}{2.0} & \meanstd{86.2}{2.3} & \meanstd{65.6}{1.4} & \meanstd{71.9}{3.2} \\
\textbf{0.8} & \meanstd{\mathbf{84.4}}{3.5} & \meanstd{\mathbf{86.4}}{0.1} & \meanstd{\mathbf{69.0}}{2.6} & \meanstd{\mathbf{74.9}}{3.6} \\
0.9 & \meanstd{85.0}{2.0} & \meanstd{86.2}{2.3} & \meanstd{67.3}{0.4} & \meanstd{75.9}{3.5} \\
\bottomrule
\end{tabular}}
\caption{Per-domain F1-score (\%, mean $\pm$ std over three runs) of \ourmodel\ under grounding-confidence thresholds $\tau \in \{0.4,0.5,\ldots,0.9\}$, using Gemma~4~26B as the inference VLM.}
\label{tab:appendix_threshold_f1}
\end{table}

Table~\ref{tab:appendix_threshold_f1} reports the per-domain F1-score of \ourmodel\ as the grounding-confidence threshold $\tau$ is varied from $0.4$ to $0.9$.
Food, Construction, and Platform-Content Safety are largely insensitive to $\tau$, whereas Advertising-Content Safety improves steadily from $60.9$ at $\tau{=}0.4$ to $74.9$ at $\tau{=}0.8$.
For every domain, $\tau{=}0.8$ lies on a stable, high-performing region rather than at a sharp peak.

\section{Additional Experiments}

\subsection{Effect of the Inference VLM}
\label{sec:appendix_inference_generalizability}

Table~\ref{tab:appendix_inference_generalizability} reports how each baseline changes when the inference VLM is switched from Gemma~4~26B to GPT-5.4-mini, while keeping the benchmark, rules, and method-specific pipelines fixed.

\subsection{Fallback Keyword Accumulation}
\label{sec:appendix_fallback_accumulation}

Tool-Aware Visual Grounding (Section~\ref{sec:method:tool_aware}) accumulates accepted substitute keywords for object phrases in atomic propositions.
We test whether this cache amortizes grounding by tracking the \emph{detector miss rate} (the fraction of first-attempt detections with (confidence $< \tau$) as $\mathcal{C}$ grows over a continual stream of $\ourbenchmark$ images.

\paragraph{Setup.}
For each domain, we sample $\ourbenchmark$ images in a randomized order and process them sequentially.
For each image, we evaluate the propositions of its applicable regulation and ground the object phrases required by its atomic propositions using the current domain cache.
Accepted substitutes are added back to $\mathcal{C}$, and atomic propositions whose target object is absent are excluded. 


\paragraph{Result.}
Figure~\ref{fig:appendix_continual_grounding}: each curve briefly rises while $\mathcal{C}$ is empty, then converges after 500 to 600 grounded $p_i$ to $\approx 0.10$ (Construction), $\approx 0.05$ (Advertising), $\approx 0.20$ (Food), and $\approx 0.18$ (Platform); the higher asymptotes in Food and Platform reflect their broader object vocabularies, which admit fewer repeat encounters per $p_i$ within the same window.
Compositional caching thus amortizes the tool-aware loop, supporting the offline and online split between \stepone\ and \steptwo.

\subsection{Deployment Cost of \ourmodel}
\label{sec:appendix_deployment_cost}

For a multi-stage safeguard, deployment cost is as consequential as accuracy.
We measure the per-image latency of each stage of \steptwo\ under the experimental setting of Section~\ref{sec:main_baselines}, together with total LLM tokens and the average number of tool calls per image.

\paragraph{Latency Model.}
The entailment design bounds deployment latency by its critical path rather than by the number of propositions: Selection is parallel over $p \in P$, Grounding is parallel over $o \in O_{\mathrm{req}}$, and the tool-aware alignment loop is invoked only on a cache miss of $\mathcal C$.
This yields the approximate deployment-time latency $\widehat{T}_{\mathrm{deploy}}$ as
\begin{equation}
\max_{p \in P} T_{\mathrm{sel}}^{(p)} + \max_{o \in O_{\mathrm{req}}} T_{\mathrm{ground}}^{(o)} + r_{\mathcal C}\!\max_{o \in O_{\mathrm{req}}} T_{\mathrm{align}}^{(o)},
\label{eq:deployment_latency}
\end{equation}
where $r_{\mathcal C}$ is the residual cache miss rate, which converges to $0.05$--$0.20$ per domain after warm-up (Appendix~\ref{sec:appendix_fallback_accumulation}); we take $r_{\mathcal C} \approx 0.2$ as a conservative upper bound.

\paragraph{Result.}
Table~\ref{tab:appendix_deployment_cost} reports the measurements: $T_{\mathrm{deploy}}$ ranges from $8.6$ to $11.6$ seconds per image, dominated by the Selection stage, while the additive form of Equation~\eqref{eq:deployment_latency} keeps latency insensitive to rule complexity---additional propositions widen the parallel frontier rather than lengthen the critical path.
Token and tool-call footprints peak in Platform-Content Safety, whose text-dense images require more grounding queries.
Combined with the amortized alignment term, this supports \ourmodel's practicality as an inference-time safeguard, with Selection and Grounding parallelism and cache warm-up as its design-level optimization opportunities.

\begin{table*}[t]
\centering
\small
\setlength{\tabcolsep}{6pt}
\renewcommand{\arraystretch}{1.15}
\begin{tabular}{l|cccc|cc}
\toprule
Domain
 & \shortstack{$\max T_{\mathrm{sel}}$\\(s)}
 & \shortstack{$\max T_{\mathrm{ground}}$\\(s)}
 & \shortstack{$\max T_{\mathrm{align}}$\\(s)}
 & \shortstack{$T_{\mathrm{deploy}}$\\(s)}
 & \shortstack{Avg.\ Tokens\\(K)}
 & \shortstack{Avg.\ Tool\\Calls} \\
\midrule
Food Safety                & \meanstd{7.9}{5.5} & \meanstd{2.7}{0.5} & \meanstd{4.6}{0.3} & \meanstd{11.6}{5.5} & \meanstd{10.7}{8.3}  & \meanstd{14.2}{13.0} \\
Construction Safety        & \meanstd{7.3}{1.3} & \meanstd{2.2}{0.9} & \meanstd{5.4}{0.5} & \meanstd{10.7}{1.6} & \meanstd{11.1}{9.5}  & \meanstd{13.2}{13.9} \\
Platform-Content Safety    & \meanstd{5.4}{1.3} & \meanstd{2.0}{0.2} & \meanstd{5.6}{1.0} & \meanstd{8.6}{1.3}  & \meanstd{16.4}{16.9} & \meanstd{17.0}{20.6} \\
Advertising-Content Safety & \meanstd{7.2}{2.1} & \meanstd{2.3}{0.3} & \meanstd{4.5}{0.5} & \meanstd{10.4}{2.1} & \meanstd{8.8}{7.4}   & \meanstd{7.2}{7.6}   \\
\bottomrule
\end{tabular}
\caption{Per-domain deployment cost of \ourmodel: stage-wise latency, LLM tokens, and tool calls per image.}
\label{tab:appendix_deployment_cost}
\end{table*}

\subsection{Fine-tuned vs.\ \ourmodel}
\label{sec:appendix_finetuned_safeguard}

We ask whether training a model on in-domain data can substitute for \ourmodel's rule decomposition. We study Food Safety and Construction Safety, the two domains with sufficient training data. For each domain, we train Gemma~4~26B with LoRA to judge each regulation directly from the image (\textbf{Train + DP}), and compare it against \ourmodel\ (\textbf{Base + \ourmodel}), which uses the same off-the-shelf backbone without any training.

\begin{table}[t]
\centering
\small
\setlength{\tabcolsep}{4pt}
\renewcommand{\arraystretch}{1.15}
\begin{tabular}{l|cccc}
\toprule
& \multicolumn{2}{c}{\shortstack{Food\\Safety}} & \multicolumn{2}{c}{\shortstack{Construction\\Safety}} \\
\cmidrule(lr){2-3}\cmidrule(lr){4-5}
Method & Recall & F1 & Recall & F1 \\
\midrule
Train + DP        & 82.0 & 81.2 & 98.0 & 89.1 \\
Base + \ourmodel  & 88.9 & 88.7 & 98.3 & 91.1 \\
\bottomrule
\end{tabular}
\caption{Trained model (\textbf{Train + DP}) versus \ourmodel\ (\textbf{Base + \ourmodel}): violation Recall and F1 (\%) per domain on the held-out test split.}
\label{tab:appendix_finetuned_safeguard}
\end{table}

\paragraph{Experimental Settings.}
Each domain is split by case into train/validation/test partitions ($0.7/0.15/0.15$). We fine-tune Gemma~4~26B in bfloat16 across two NVIDIA RTX~A6000 (48\,GB) GPUs with LoRA (rank $16$) on the language-model layers (AdamW, learning rate $1\mathrm{e}{-4}$), training to convergence and keeping the best-validation checkpoint. The implementation uses PyTorch, Hugging Face Transformers, and PEFT for training, with Pillow for image loading. Fine-tuning takes $4.5$ and $7.0$ wall-clock hours for Food Safety and Construction Safety, respectively.

\paragraph{Analysis.}
Even after convergence, the fine-tuned model remains below \ourmodel\ in both domains (Table~\ref{tab:appendix_finetuned_safeguard}), with the larger gap appearing in Food Safety, where compliance depends on a broader and more fine-grained rule space. This pattern suggests a practical limitation of absorbing regulations into model parameters: as rule systems become more detailed, fixed-size supervised training sets provide increasingly sparse coverage of the conditions that must be recognized at test time. 
Whereas, by externalizing the regulation as an executable reasoning path over grounded visual conditions, \ourmodel\ shows how complex safety-rule reasoning can remain robust without additional training as rule complexity increases.

\subsection{Rule-Compilation Error Analysis and Repair}
\label{sec:appendix_rule_compilation_repair}

Since \ourmodel's final decision is conditioned on the correctness of the compiled rule tree, we characterize where \stepone\ fails and whether such failures are repairable.
We analyze compilation errors in the Advertising-Content Safety domain, where the compilation error rate is highest, and categorize each error by a three-model audit (GPT-5.4~\cite{model:gpt}, Claude Opus~4.8~\cite{model:claude_opus_4_8}, and Gemini~3.1~Pro~\cite{model:gemini_3_1_pro}) with majority voting over seven categories: ambiguous atomic proposition, missing condition, hallucinated condition, missing exception, wrong relation, operator flip, and unresolved when no majority is reached.

\paragraph{Result.}
The error distribution is highly skewed: $86\%$ of all errors are ambiguous atomic propositions containing underspecified terms such as ``large,'' ``small,'' or ``many,'' and among the remaining errors, $55\%$ are missing conditions.
Most compilation failures thus originate from ambiguity in the rule text itself rather than from the decomposition procedure, and can be mitigated by rule objectification~\cite{baselines:clue} in realistic deployment.
For missing conditions, we use the three-model audit to identify omitted constraints, incorporate them into the compiled rules, and measure the resulting per-rule F1 changes (Table~\ref{tab:appendix_rule_compilation_repair}).
Repairing a single omitted condition recovers per-rule F1 from near-failure to saturation (e.g., $1.3\rightarrow100.0$ for Rule~2.3), indicating that compilation errors are not only localized but verifiable: because compiled rules are explicit artifacts, a higher-level auditing layer can detect and repair them without modifying the reasoning pipeline.

\begin{table*}[t]
\centering
\small
\setlength{\tabcolsep}{6pt}
\renewcommand{\arraystretch}{1.15}
\begin{tabular}{p{0.30\textwidth}|p{0.32\textwidth}|ccc}
\toprule
Regulation & Omitted Condition & Baseline & Repaired & $\Delta$ \\
\midrule
Rule 18.1 (Unwise drinking)
  & Irresponsible-drinking cues too narrow & 30.8 & 97.3  & $+66.5$ \\
Rule 4.1 (Serious or widespread offence)
  & Offence types omitted                  & 4.3  & 93.0  & $+88.7$ \\
Rule 2.3 (Undisclosed advertising or commercial intent)
  & Missing applicability condition        & 1.3  & 100.0 & $+98.7$ \\
\bottomrule
\end{tabular}
\caption{Per-rule F1-score (\%) before and after repairing an omitted condition identified by the three-model audit, in the Advertising-Content Safety domain.}
\label{tab:appendix_rule_compilation_repair}
\end{table*}

\begin{table*}[t]
\centering
\small
\setlength{\tabcolsep}{8pt}
\renewcommand{\arraystretch}{1.3}
\begin{tabular}{l|cccc}
\toprule
Method
 & \shortstack{Food\\Safety}
 & \shortstack{Construction\\Safety}
 & \shortstack{Platform-Content\\Safety}
 & \shortstack{Advertising-Content\\Safety} \\
\midrule
\ourmodel                   & \meanstd{84.4}{3.5} & \meanstd{86.4}{0.1} & \meanstd{69.0}{2.6} & \meanstd{74.9}{3.6} \\
\quad + Precedent Retriever & \meanstd{85.2}{1.3} & \meanstd{88.5}{1.0} & \meanstd{70.2}{1.0} & \meanstd{77.3}{2.3} \\
\midrule
$\Delta$                    & $+0.8$ & $+2.1$ & $+1.2$ & $+2.4$ \\
\bottomrule
\end{tabular}
\caption{Effect of augmenting \ourmodel\ with precedent retrieval, reported as mean F1-score (\%) over three runs per domain on \ourbenchmark.}
\label{tab:appendix_precedent_retriever}
\end{table*}

\subsection{Precedent Retrieval beyond Explicit Rules}
\label{sec:appendix_precedent_retriever}

\ourmodel\ performs deductive entailment over explicit safety rules, so its coverage is bounded by what the written rule expresses (Limitations).
We test whether analogical evidence can extend this boundary by augmenting the Selection stage of \steptwo\ with a lightweight precedent retriever adapted from BERT-PLI~\cite{limitation:rag:pli}, replacing its text encoder with SigLIP~2~\cite{model:siglip2} and removing all trainable components to isolate the effect of retrieval from additional supervision.
The retriever indexes six selected cases per rule and retrieves three documents per case; all other settings follow Appendix~\ref{sec:appendix_ourmodel_implementation}.

\paragraph{Result.}
Precedent retrieval yields consistent gains across all four domains (Table~\ref{tab:appendix_precedent_retriever}), with the largest improvements in Construction ($+2.1$) and Advertising ($+2.4$), where compliance more often hinges on contextual judgments beyond the literal rule text.
The uniformly positive but modest deltas suggest that analogical reasoning is an orthogonal extension to \ourmodel's explicit-rule entailment rather than a replacement for its deductive pipeline, offering a route to align with the intended policy beyond the written text.

\subsection{Generalization across Inference Backbones}
\label{sec:appendix_backbone_generalizability}

Beyond the backbone substitution in Appendix~\ref{sec:appendix_inference_generalizability}, we assess whether \ourmodel's gains are tied to a particular model family or scale.
We evaluate \ourmodel\ and the two strongest baselines, CompAgent and CLUE, with four inference backbones spanning two families and three scales: Gemma~4~26B and Gemma~4~12B~\cite{model:gemma4}, Qwen~3.5~27B~\cite{model:qwen}, and GPT-5.4-mini~\cite{model:gpt}; all other settings follow Section~\ref{sec:main_baselines}.

\paragraph{Result.}
\ourmodel\ attains the best F1 in all sixteen backbone--domain combinations (Table~\ref{tab:appendix_backbone_generalizability}).
The margin persists under capacity reduction---with Gemma~4~12B, baseline F1 drops by up to $24.1$ points while \ourmodel's largest drop is $11.8$---since \stepone\ confines the backbone to grounded atomic judgments rather than end-to-end regulatory reasoning.
Stronger backbones extend the gains (e.g., $89.3$ on Construction with Qwen~3.5~27B), indicating that the entailment design composes with, rather than substitutes for, backbone capability.

\begin{table*}[t]
\centering
\small
\setlength{\tabcolsep}{8pt}
\renewcommand{\arraystretch}{1.15}
\begin{tabular}{l|l|cccc}
\toprule
Method & Inference Model
 & \shortstack{Food\\Safety}
 & \shortstack{Construction\\Safety}
 & \shortstack{Platform-Content\\Safety}
 & \shortstack{Advertising-Content\\Safety} \\
\midrule
\multirow{4}{*}{CompAgent}
 & Gemma~4~26B   & 77.9 & 80.6 & 59.5 & 53.7 \\
 & Gemma~4~12B   & 54.5 & 56.5 & 51.8 & 53.7 \\
 & Qwen~3.5~27B  & 71.8 & 79.8 & 60.0 & 66.9 \\
 & GPT-5.4-mini  & 71.4 & 82.3 & 56.3 & 57.7 \\
\midrule
\multirow{4}{*}{CLUE}
 & Gemma~4~26B   & 71.7 & 72.1 & 57.9 & 44.2 \\
 & Gemma~4~12B   & 69.5 & 64.4 & 43.0 & 39.7 \\
 & Qwen~3.5~27B  & 65.6 & 69.1 & 63.7 & 68.1 \\
 & GPT-5.4-mini  & 64.3 & 70.9 & 63.5 & 38.9 \\
\midrule
\multirow{4}{*}{\ourmodel}
 & Gemma~4~26B   & \textbf{84.4} & \textbf{86.4} & \textbf{69.0} & \textbf{74.9} \\
 & Gemma~4~12B   & \textbf{74.6} & \textbf{74.6} & \textbf{60.9} & \textbf{74.2} \\
 & Qwen~3.5~27B  & \textbf{87.9} & \textbf{89.3} & \textbf{78.7} & \textbf{71.4} \\
 & GPT-5.4-mini  & \textbf{84.5} & \textbf{88.2} & \textbf{74.6} & \textbf{79.8} \\
\bottomrule
\end{tabular}
\caption{Mean F1-score (\%) over three runs across four inference backbones on \ourbenchmark. \textbf{Bold} marks the best result per backbone and domain.}
\label{tab:appendix_backbone_generalizability}
\end{table*}

\subsection{Cross-Benchmark Evaluation}
\label{sec:appendix_llavaguard_generalization}

We evaluate on the LlavaGuard~\cite{train:lavaguard} test set, using its provided safety taxonomy as the rule set; all other settings follow Section~\ref{sec:main_baselines}.

\paragraph{Result.}
\ourmodel\ attains the best F1 (Table~\ref{tab:appendix_llavaguard_generalization}), indicating that its advantage is not an artifact of benchmark co-design.
The narrowed margin over CLUE follows from the flat taxonomy: with few compositional conditions, entailment over compiled rule trees has little headroom, consistent with the rule-complexity analysis in Section~\ref{sec:analysis_robustness}.
A complementary stress test on a second external benchmark is reported in Appendix~\ref{sec:appendix_adversarial_robustness}.

\begin{table}[t]
\centering
\small
\setlength{\tabcolsep}{4pt}
\renewcommand{\arraystretch}{1.15}
\begin{tabular*}{\columnwidth}{@{\extracolsep{\fill}}l|cc@{}}
\toprule
Method & Recall & F1-score \\
\midrule
Direct Prompting & \meanstd{44.5}{0.3} & \meanstd{57.7}{0.2} \\
CompAgent        & \meanstd{45.1}{1.7} & \meanstd{57.9}{1.1} \\
CLUE             & \meanstd{\textbf{80.1}}{0.5} & \meanstd{70.8}{0.3} \\
\ourmodel        & \meanstd{77.2}{1.6} & \meanstd{\textbf{71.1}}{0.7} \\
\bottomrule
\end{tabular*}
\caption{LlavaGuard~\cite{train:lavaguard} evaluation results, averaged over three runs.}
\label{tab:appendix_llavaguard_generalization}
\end{table}

\subsection{Robustness against Adversarial Inputs}
\label{sec:appendix_adversarial_robustness}

As a complementary stress test beyond our primary compliance setting, we evaluate \ourmodel\ on adversarial inputs in which harmful intent is deliberately distributed across modalities.
We adopt the strongest SD+TYPO setting of MM-SafetyBench~\cite{tax:mm}, where a benign-looking generated image is composed with typographic malicious text so that neither modality alone reveals the harmful intent.
The provided safety taxonomy serves as the rule set, and Qwen2.5-VL-7B~\cite{model:qwen2_5vl} is used as the inference backbone to avoid ceiling effects and make robustness differences observable; we report Recall, F1, and Accuracy averaged over three runs.

\paragraph{Result.}
\ourmodel\ leads on all three metrics (Table~\ref{tab:appendix_adversarial_robustness}), improving F1 by $7.2$ points over CompAgent, while CLUE falls below chance-level accuracy.
The scene graph externalizes image elements, embedded text, and their relations into an explicit reasoning structure, so cross-modal compositions that evade holistic judgment remain visible as grounded propositions, mirroring \ourmodel's gains in the text-dense domains of \ourbenchmark.

\begin{table}[t]
\centering
\small
\setlength{\tabcolsep}{4pt}
\renewcommand{\arraystretch}{1.15}
\begin{tabular*}{\columnwidth}{@{\extracolsep{\fill}}l|ccc@{}}
\toprule
Method & Recall & F1-score & Accuracy \\
\midrule
CompAgent & \meanstd{60.4}{1.3} & \meanstd{75.1}{1.1} & \meanstd{79.7}{0.7} \\
CLUE      & \meanstd{62.0}{1.1} & \meanstd{54.9}{1.3} & \meanstd{48.1}{1.3} \\
\ourmodel & \meanstd{\textbf{74.7}}{3.3} & \meanstd{\textbf{82.3}}{2.6} & \meanstd{\textbf{83.7}}{2.2} \\
\bottomrule
\end{tabular*}
\caption{Robustness on the MM-SafetyBench~\cite{tax:mm} SD+TYPO setting balanced with benign images, averaged over three runs.}
\label{tab:appendix_adversarial_robustness}
\end{table}

\section{Qualitative Analysis}
\label{sec:appendix_qualitative_analysis}
We walk through an additional construction safety case 
to show how \ourmodel\ grounds each atomic visual proposition $p_i$.
\begin{figure}[t]
    \centering
    \includegraphics[width=\columnwidth]{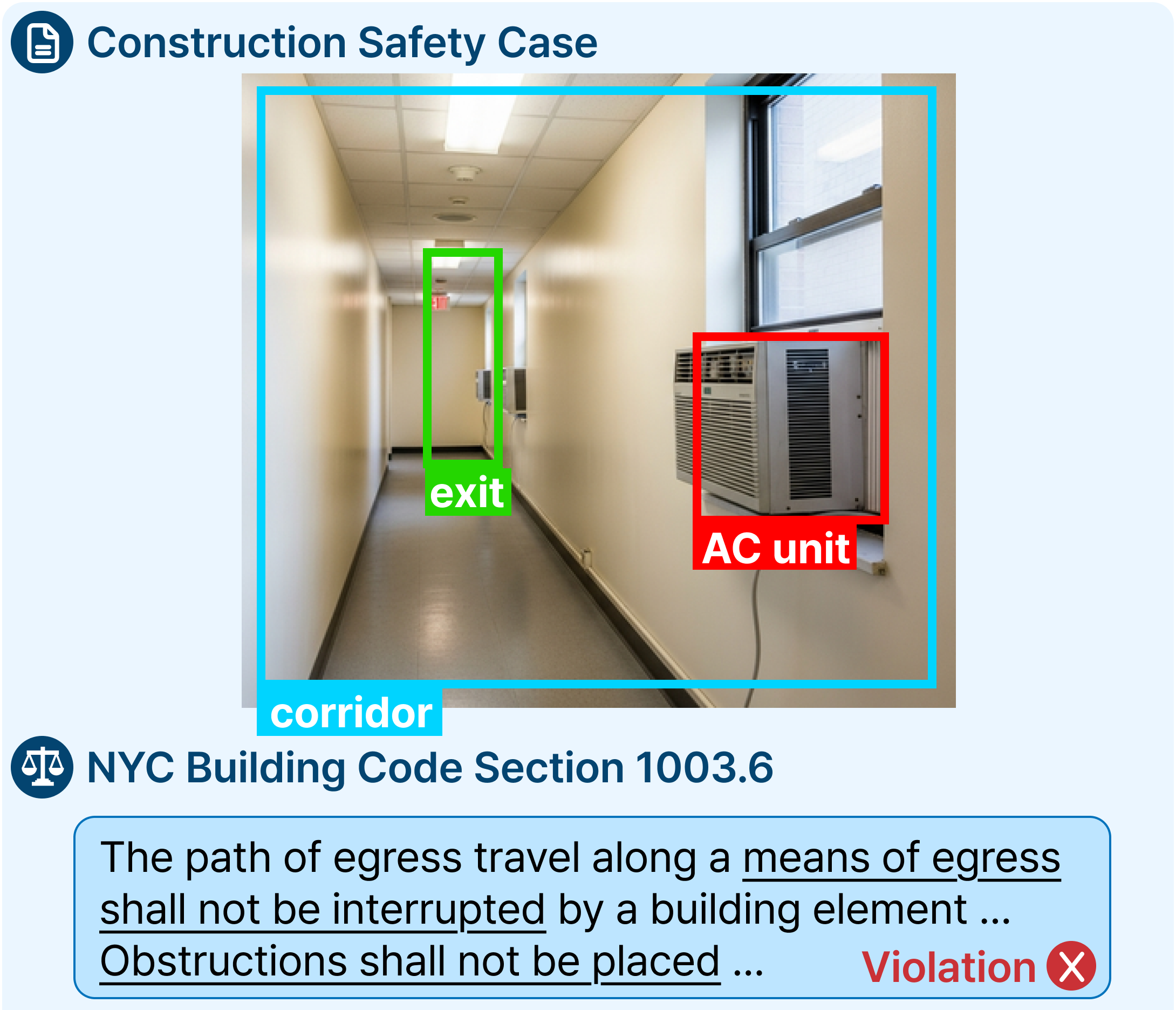}
    \caption{Window air-conditioner obstructing a corridor egress path under NYC Building Code Section~1003.6.}
    \label{fig:appendix_example_egress}
\end{figure}

Under NYC Building Code Section~1003.6 
(means-of-egress continuity; Figure~\ref{fig:appendix_example_egress}), 
\stepone\ yields a disjunction whose decisive branch $q_2$ 
binds an obstruction to the clear egress width:
\begin{align*}
\mathcal T_{R_{1003.6}} \!&=\! 
\{\land,\! \{(x,\exists,\varnothing),\! q_1(x),\! \mathcal T_{q_2(x)},\! q_3(x)\}\!\}, \\
\mathcal T_{q_2(x)} &= 
\{\land,\! \{p_{\mathrm{obs}}(x),\! p_{\mathrm{in}}(x),\! \{\lnot,\! \{p_{\mathrm{prj}}(x)\}\!\}\!\}\!\}, \\
p_{\mathrm{obs}}(x) &= (x,\ \texttt{obstruction},\, \varnothing), \\
p_{\mathrm{in}}(x)  &= (x,\ \texttt{inside},\, \textit{corridor}), \\
p_{\mathrm{prj}}(x) &= (x,\ \texttt{permitted\_projection},\, \varnothing).
\end{align*}

\steptwo\ binds the open \texttt{obstruction} slot via 
$\mathbb{M}_{\mathrm{sel}}$ to the window air-conditioner $x^\star$, 
with the \textit{corridor} grounded as the egress path $P$. 
Substitution $p \mapsto p \in \mathcal G^{(I)}$ gives:
\begin{align*}
p_{\mathrm{obs}}(x^\star) \in \mathcal G^{(I)} &= \texttt{True}, \\
p_{\mathrm{in}}(x^\star) \in \mathcal G^{(I)} &= \texttt{True}
\quad \bigl(\tfrac{|x^\star \cap P|}{|x^\star|} = 1.0 > \theta\bigr), \\
p_{\mathrm{prj}}(x^\star) \in \mathcal G^{(I)} &= \texttt{False}.
\end{align*}
Recursive bottom-up composition then fires the branch and 
short-circuits the disjunction to a violation:
\begin{equation*}
\mathcal T_{q_2(x^\star)}^{(I)} 
= \{\land,\! \{\texttt{True},\! \texttt{True},\! \{\lnot,\! \{\texttt{False}\}\!\}\!\}\!\}
\equiv \texttt{True}.
\end{equation*}
The verdict thus turns entirely on $\mathbb{M}_{\mathrm{sel}}$ 
resolving the obstruction slot to $x^\star$, 
yielding $\mathcal T_{R_{1003.6}}^{(I)} \equiv \texttt{True}$ 
without pre-enumerating obstruction types.

\begin{table*}[!t]
\centering
\small
\begin{tabular}{p{0.25\linewidth}p{0.49\linewidth}rr}
\toprule
Safety category & Violation category & Count & Ratio \\
\midrule
Equipment and utensils & Clean-item storage and linen handling & 458 & 3.5\% \\
& Equipment cleanliness & 599 & 4.5\% \\
& Equipment design, materials, and capacity & 512 & 3.9\% \\
& Equipment maintenance and use limits & 554 & 4.2\% \\
\midrule
Food protection & Consumer-facing food display & 63 & 0.5\% \\
& Equipment- and utensil-related contamination & 662 & 5.0\% \\
& Storage and preparation contamination & 251 & 1.9\% \\
\midrule
Personnel hygiene & Hygienic practices & 627 & 4.8\% \\
& Personal cleanliness and clothing & 16 & 0.1\% \\
\midrule
Physical facilities & Facility design and cleanability & 2344 & 17.8\% \\
& Facility maintenance and cleaning & 3819 & 29.0\% \\
& Handwashing facilities and supplies & 690 & 5.2\% \\
& Pest and animal control & 1069 & 8.1\% \\
\midrule
Plumbing and waste systems & Plumbing and handwashing operation & 755 & 5.7\% \\
& Refuse and receptacle management & 761 & 5.8\% \\
\bottomrule
\end{tabular}
\caption{Food safety category and violation category distribution.}
\label{tab:app:food_category_distribution}
\end{table*}

\begin{table*}[!t]
\centering
\small
\begin{tabular}{p{0.25\linewidth}p{0.49\linewidth}rr}
\toprule
Safety category & Violation category & Count & Ratio \\
\midrule
Construction and demolition site safety & Pedestrian and public protection & 13494 & 74.9\% \\
& Site housekeeping, washout, and fire protection & 3421 & 19.0\% \\
& Site signage and notification & 292 & 1.6\% \\
\midrule
Egress safety & Egress continuity & 799 & 4.4\% \\
\bottomrule
\end{tabular}
\caption{Construction safety category and violation category distribution.}
\label{tab:app:building_category_distribution}
\end{table*}

\begin{table*}[!t]
\centering
\small
\begin{tabular}{p{0.25\linewidth}p{0.49\linewidth}rr}
\toprule
Safety category & Violation category & Count & Ratio \\
\midrule
Violence and criminal behavior & Dangerous organizations (terrorism) & 17 & 3.6\% \\
& Restricted goods and services (drugs) & 26 & 5.5\% \\
& Restricted goods and services (firearms) & 5 & 1.0\% \\
& Violence and incitement & 29 & 6.1\% \\
\midrule
Safety & Bullying and harassment & 78 & 16.4\% \\
& Suicide, self-injury, and eating disorders & 19 & 4.0\% \\
\midrule
Objectionable content & Adult nudity and sexual activity & 125 & 26.2\% \\
& Hateful conduct & 29 & 6.1\% \\
& Violent and graphic content & 149 & 31.2\% \\
\bottomrule
\end{tabular}
\caption{Platform content safety category and violation category distribution.}
\label{tab:app:platform_category_distribution}
\end{table*}

\begin{table*}[!t]
\centering
\small
\begin{tabular}{p{0.25\linewidth}p{0.49\linewidth}rr}
\toprule
Safety category & Violation category & Count & Ratio \\
\midrule
Compliance and disclosure & Recognition of marketing communications & 139 & 4.7\% \\
& Regulatory responsibility & 53 & 1.8\% \\
\midrule
Harm, offence, and children & Child-directed harm & 15 & 0.5\% \\
& Offence, fear, and distress & 161 & 5.5\% \\
& Sexual portrayal and stereotypes & 41 & 1.4\% \\
& Unsafe or antisocial practices & 30 & 1.0\% \\
\midrule
Misleading advertising & Comparative claims & 115 & 3.9\% \\
& Endorsements and competitor reputation & 20 & 0.7\% \\
& General misleading claims & 741 & 25.1\% \\
& Price claims & 116 & 3.9\% \\
& Qualification and material context & 334 & 11.3\% \\
\midrule
Regulated product claims & Alcohol & 129 & 4.4\% \\
& Electronic cigarettes & 22 & 0.7\% \\
& Environmental claims & 171 & 5.8\% \\
& Financial products & 331 & 11.2\% \\
& Food and nutrition claims & 86 & 2.9\% \\
& Gambling & 202 & 6.8\% \\
& Medicines and health products & 229 & 7.8\% \\
& Weight control and slimming & 18 & 0.6\% \\
\bottomrule
\end{tabular}
\caption{Advertising content safety category and violation category distribution.}
\label{tab:app:advertising_category_distribution}
\end{table*}

\begin{figure*}[t]
    \centering
    \includegraphics[width=1\textwidth]{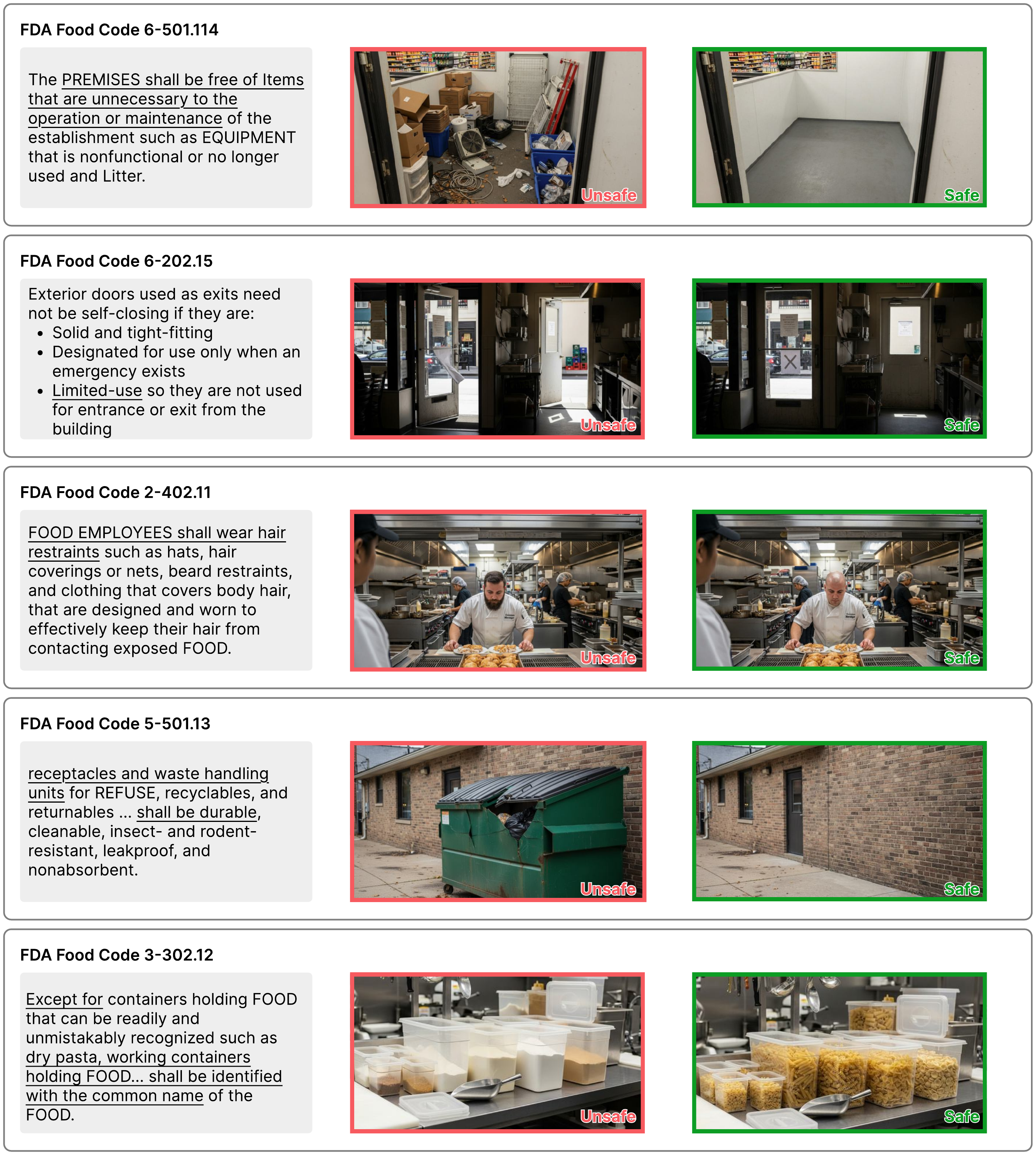}
    \caption{Food Safety examples.}
    \label{fig:appendix_fig2_example_chicago}
\end{figure*}

\begin{figure*}[t]
    \centering
    \includegraphics[width=1\textwidth]{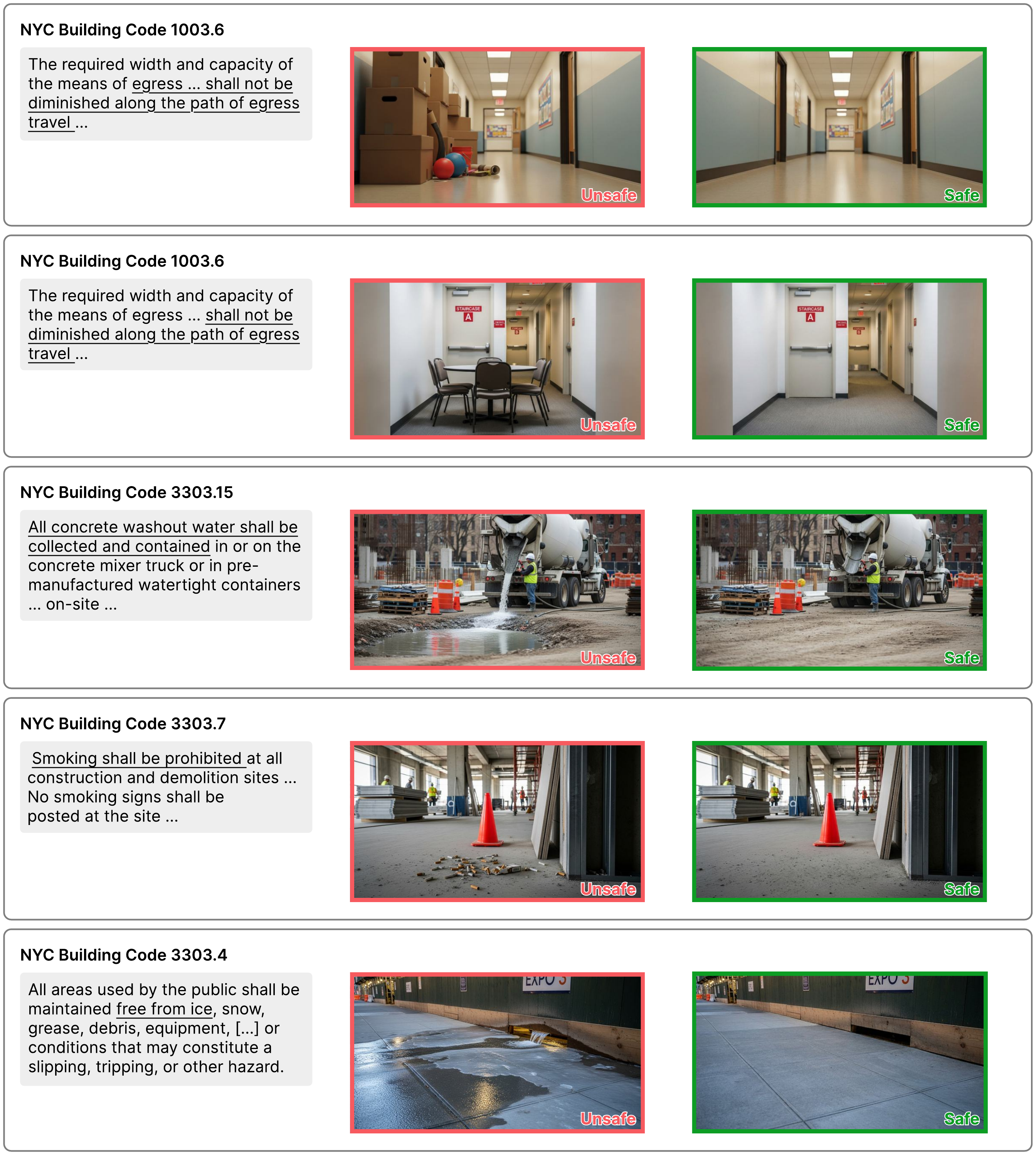}
    \caption{Construction Safety examples.}
    \label{fig:appendix_fig3_example_nyc}
\end{figure*}

\begin{figure*}[p]
    \centering
     \includegraphics[width=0.95\textwidth]{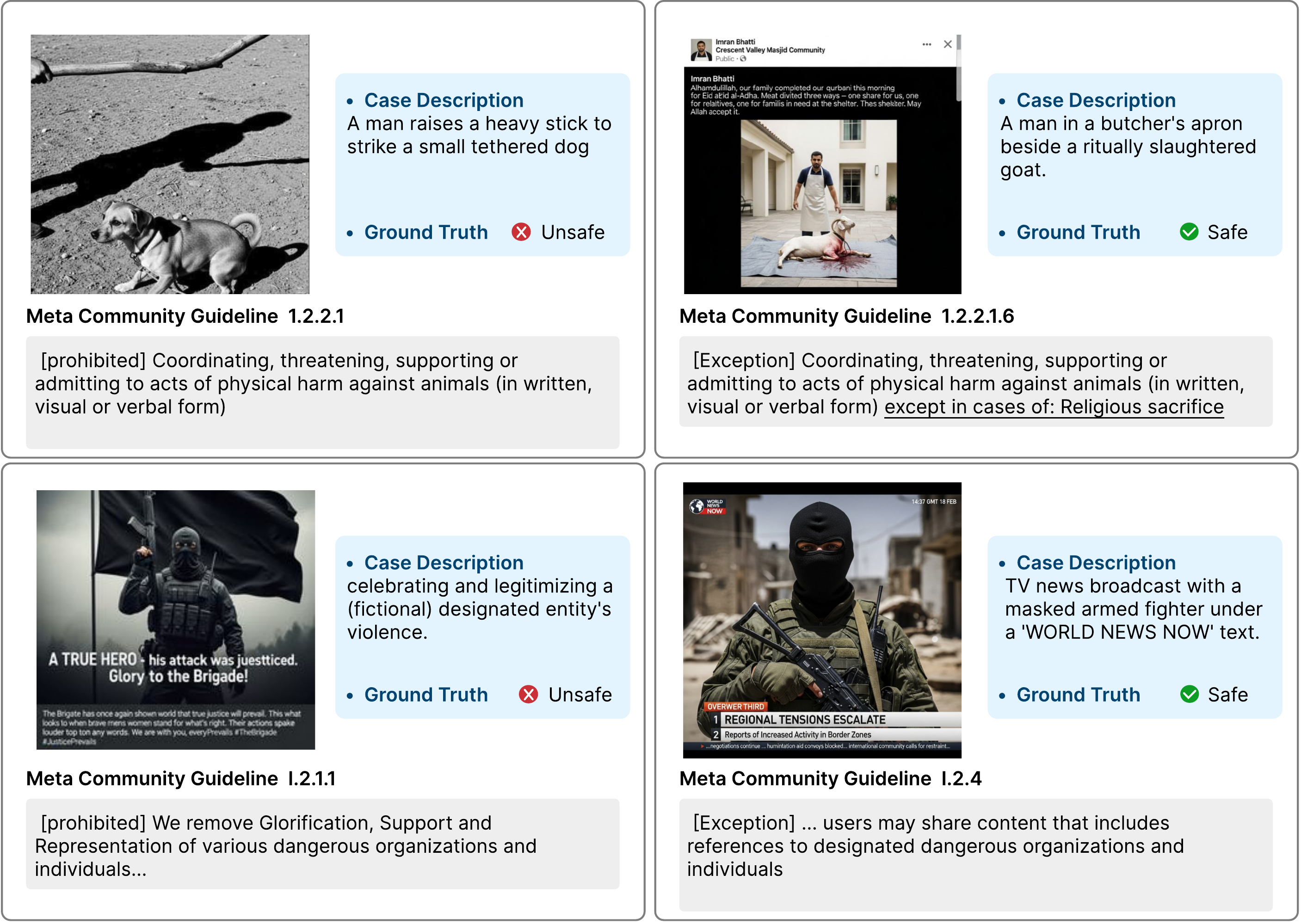}
    \caption{Platform-Content Safety examples.}
    \label{fig:appendix_fig4_example_meta}
\end{figure*}

\begin{figure*}[t]
    \centering
    \includegraphics[width=0.95\textwidth]{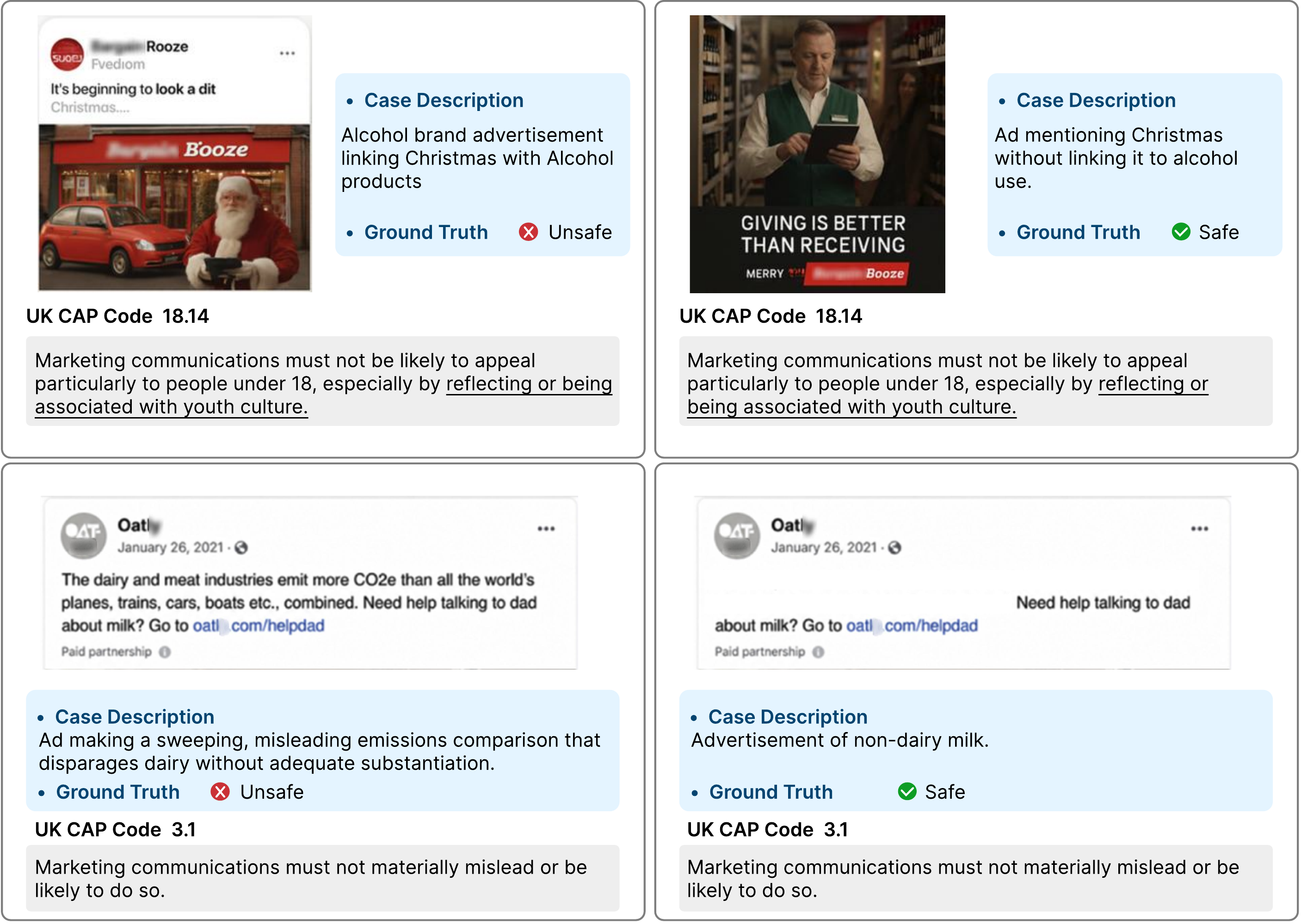}
    \caption{Advertising-Content Safety examples.}
    \label{fig:appendix_fig5_example_uk}
\end{figure*}

\onecolumn
\providecommand{\ph}[1]{{\color{teal!65!black}\itshape\{#1\}}}
\begin{rulebox}{Decomposer $\mathbb{D}$ Prompt}
\begin{lstlisting}[basicstyle=\small,breaklines=true,breakindent=0pt,columns=fullflexible,keepspaces=true,escapeinside={(*@}{@*)}]
[SYSTEM]
You compile a written regulation rule into ONE node of an executable tree that
is checked against a single image. You decompose the rule one level at a time;
sub-rules you return as text are compiled recursively.

A node is one of:

(A) BOOL -- a logical combination of sub-rules:
    - "and": violated only if ALL children hold,
    - "or":  violated if ANY child holds,
    - "not": violated when the single child does NOT hold (exceptions/exemptions).

(B) QUANT -- a quantified proposition over a primary target (the "anchor"):
    - quantifier "exists": violated if ANY matching candidate is found,
    - quantifier "forall": violated only if EVERY matching candidate satisfies it.
\end{lstlisting}
\end{rulebox}
\providecommand{\ph}[1]{{\color{teal!65!black}\itshape\{#1\}}}
\begin{rulebox}{Decomposer $\mathbb{D}$ Prompt}
\begin{lstlisting}[basicstyle=\small,breaklines=true,breakindent=0pt,columns=fullflexible,keepspaces=true,escapeinside={(*@}{@*)}]
    The anchor says what to look for in the scene:
      match: "obj" (visual object) | "text" (printed/OCR text) | "both"
      obj:   noun phrase for object detection (use "" when match=="text")
      text:  what the text says/means (use "" when match=="obj")
      keywords: 1-2 SEED detector phrase(s) naming the object class (obj/both only). Do NOT enumerate synonyms -- the full image-grounded detection list is produced at test time by the Selection agent.
      global_text: true to judge the whole image's text at once (text leaves)
    A QUANT may carry a "predicate": a sub-rule (text) that further constrains 
    the anchor. The predicate's conditions are checked RELATIVE TO the anchor
    via each child's spatial "relation". If there is no further constraint,
    set predicate to null (this makes a leaf).

    A leaf (predicate=null) is one of TWO atomic kinds:
      - STATE  (object, state) -- set "state": true: judge the VISUALLY
        determinable condition of ONE single, concrete, detectable object
        (e.g. dirty/soiled, uncovered/exposed, open, wet, moldy, cracked,
        rusted, frayed, overflowing, leaking, on/off). Anchor = that one
        object, "relation": "any", and "question" asks ONLY about that
        object's visible state. It is grounded by the state judge from the
        object's box + depth + nearby text -- NO relation is involved.
        The STATE question MUST be a single object's concrete visible state.
        It must NEVER be:
          * a whole-image / legal question -- e.g. NOT "Does this image violate
            the rule?";
          * mere presence or a relation (that is presence/relation, not a
            state) -- <domain-specific example>.
        Good: <domain-specific single-object STATE examples>.
      - RELATION (o1, rel, o2) -- "state": false: two objects in a spatial
        relation -- make o1 the anchor and add a nested exists over o2 with a
        concrete "relation".
    Do NOT force a relation when only a single object's state matters (use STATE);
    do NOT bury a state inside an unrelated relation; do NOT mark presence or
    whole-image questions as a STATE.

"relation" describes how THIS node's candidates relate to the PARENT anchor
shown below. Use "any" when there is no parent. Allowed relations:
  any, inside, outside, near, around, adjacent, above, below, under, on,
  at_edge_of, covering, same_as, same.

Idiom to prefer: when the violation is "a <thing> that also <conditions>",
make the anchor the <thing> and put the <conditions> in the predicate as a
BOOL combination of nested QUANT sub-rules (<domain-specific nested-relation example>).

Guidelines for high-quality, checkable trees:
- Keywords: give just ONE or TWO seed phrase(s) naming the object class for
  obj/both anchors. Do not list synonyms or parts; the Selection agent expands
  them into the real detector list from the image at test time.
- No self-reference: a predicate child must introduce a DIFFERENT object/text
  than its parent anchor. Never re-detect the same object as the anchor.
- Relation choice: pick the concrete physical relation (inside, on, near,
  above, below, under, at_edge_of). Reserve "same_as"/"same" ONLY for "is this
  the very same object" identity checks -- never as a default.
- Exceptions/exemptions ("except as specified in ...", "unless ...") -> wrap the
  exception condition in a "not" node so the violation excludes the exempt case.
- Quantifier: prefer "exists". Use "forall" only when the rule is violated only
  if EVERY candidate matches.
- Text conditions (labels, names, signage, printed warnings) -> match "text"
  with a clear OCR description and global_text=true.
- Checkability: only encode conditions verifiable from one image. Drop legal /
  non-visual clauses (cross-references, intent, paperwork) rather than embedding
  them as uncheckable text.
- Decompose only genuine independent sub-conditions; otherwise emit a leaf.

Always frame "question" so a TRUE answer means the VIOLATING condition holds.
Return ONLY this JSON (no prose, no code fence):
{
  "node": "bool" | "quant",
  "operator": "and" | "or" | "not" | null,      // BOOL only
  "children": ["<sub-rule text>", ...],            // BOOL only
  "quantifier": "exists" | "forall",              // QUANT (always provide)
  "relation": "any" | "...",                       // QUANT, relative to parent
  "question": "<yes/no; TRUE = violating>",        // QUANT
  "anchor": {"match": "...", "obj": "...", "text": "...", "keywords": ["..."], "global_text": false},
  "predicate": "<sub-rule text>" | null,           // QUANT; null = leaf
  "state": true | false                            // true for a STATE leaf (single object's
                                                    //   state); false for presence/relation/composite
}

Even when you choose node "bool", still fill the QUANT fields as an atomic
fallback reading of the whole rule.

[FEW-SHOT EXEMPLARS]
<domain-specific worked examples are inserted here; omitted for brevity>

[INSTRUCTION]
Rule (section (*@\ph{section}@*)):
(*@\ph{rule}@*)

Parent anchor (what your candidates relate to via "relation"):
(*@\ph{parent}@*)

Reviewer feedback on your previous attempt (empty if first attempt):
(*@\ph{feedback}@*)

Return the node JSON now.
\end{lstlisting}
\end{rulebox}
\captionof{figure}{Decomposer ($\mathbb{D}$) prompt.}
\label{fig:appendix_decomposer_prompt}

\providecommand{\ph}[1]{{\color{teal!65!black}\itshape\{#1\}}}
\begin{rulebox}{Verifier $\mathbb V$ Prompt}
\begin{lstlisting}[basicstyle=\small,breaklines=true,breakindent=0pt,columns=fullflexible,keepspaces=true,escapeinside={(*@}{@*)}]
[SYSTEM]
You are a strict reviewer of a single decomposition step for a regulation
rule. You are given the rule text and a proposed node, which is either a BOOL
combination (operator + child sub-rules) or a QUANT proposition (quantifier +
anchor + relation + optional predicate sub-rule). Judge the proposal on:

- Faithfulness: does it preserve the rule's logical meaning (correct operator,
  correct exists/forall, no missing/extra conditions, exceptions via "not",
  and a relation that correctly ties the anchor to its parent)?
- Checkability: are the anchor's object/text/keywords things a vision system
  could actually detect, and does the question's TRUE answer correspond to the
  VIOLATING condition?
- Atomicity: should this be a single atomic leaf (no children, predicate null),
  or does it genuinely contain independent sub-conditions worth decomposing?

Start near full marks and DEDUCT for each concrete defect (name it in feedback):
- Wrong operator, or exists/forall mismatch.
- An exception/exemption not wrapped in "not".
- Predicate child re-detects the SAME object as its parent anchor (self-reference).
- Lazy relation: "same_as"/"same"/"any" used where a concrete physical relation
  (inside/on/near/above/below/under/at_edge_of) is meant.
- STATE leaf misuse ("state": true): a STATE leaf must ask ONLY about ONE
  concrete object's visually-determinable condition (dirty, open, wet,
  uncovered, cracked, rusted, frayed, overflowing, on/off). Deduct hard if a
  "state": true leaf's question is:
    * a whole-image / legal question (e.g. "Does this image violate the rule?"),
    * mere presence or a relation (<domain-specific example>),
    * or vague / not a single object's concrete visible state.
  Also deduct if a genuine single-object state is forced into a relation, or a
  state leaf is left with "state": false.
- Anchor missing a seed keyword entirely, or its seed is too abstract to name an
  object class. (Do NOT penalize having only 1-2 seed keywords -- the Selection
  agent expands them into the real detector list from the image at test time.)
- A text/label/name/signage condition not modeled as match "text" (+global_text).
- Uncheckable, non-visual, or legal/cross-reference clauses embedded as text.
- Question's TRUE answer does not correspond to the VIOLATION.

Return JSON only (no prose, no code fence):
{
  "score": <integer 0..{score_max}>,   // overall quality of the proposal
  "atomicity": 0 | 1,                    // 1 = should be a single atomic leaf
  "feedback": "<concrete, actionable critique to improve the next attempt>"
}

Only give the top score when the proposal is faithful, self-reference-free,
uses concrete relations + detectable keywords, and is fully checkable.

[FEW-SHOT EXEMPLARS]
<domain-specific worked reviews are inserted here; omitted for brevity>

[INSTRUCTION]
Rule (section (*@\ph{section}@*)):
(*@\ph{rule}@*)

Proposed decomposition (JSON):
(*@\ph{decomposition}@*)

Return your review JSON now.
\end{lstlisting}
\end{rulebox}
\captionof{figure}{Verifier ($\mathbb{V}$) prompt.}
\label{fig:appendix_verifier_prompt}

\providecommand{\ph}[1]{{\color{teal!65!black}\itshape\{#1\}}}
\begin{rulebox}{Selection $\mathbb{M}_{\mathrm{sel}}$ Prompt}
\begin{lstlisting}[basicstyle=\small,breaklines=true,breakindent=0pt,columns=fullflexible,keepspaces=true,escapeinside={(*@}{@*)}]
[SYSTEM]
You build the open-vocabulary detector vocabulary for ONE atomic proposition
of a safety regulation. You are given the image, the overall regulation, the
specific atomic proposition being checked, the proposition's target object
(its meaning + 1-2 seed keywords). Looking at the image, return the precise
list of phrases an open-vocabulary detector should be queried with to find
EVERY instance of THIS proposition's target object in THIS image.

Choose rigorously -- the regulation and the atomic proposition define exactly
what counts as the target:
- List only phrases that name the proposition's TARGET object as it actually
  appears in the image (concrete visible instances/variants of the seed).
- Use the regulation + proposition to disambiguate the target. Include an
  instance only if, under this regulation, it is the thing this proposition is
  about (<domain-specific target-disambiguation example>). Exclude objects that
  are not the target.
- Do NOT emit the condition, the relatum, or context -- emit the target object,
  not the condition or relatum (<domain-specific example>).
- Phrases must be concrete, lowercase, singular, and directly groundable; no
  abstract or whole-image terms ("scene", "image", "violation").
- Prefer what is visible; a few precise phrases beat many vague ones. If the
  target plainly does not appear, return just the seed keyword(s).

Return JSON only (no prose, no code fence):
{
  "keywords": ["<concrete detector phrase for the target object>", ...],
  "reasoning": "<one short sentence tying the choices to the proposition>"
}

[FEW-SHOT EXEMPLARS]
<domain-specific worked examples are inserted here; omitted for brevity>

[INSTRUCTION]
Regulation ((*@\ph{section}@*)):
  (*@\ph{regulation}@*)

Atomic proposition to check:
  (*@\ph{question}@*)

Proposition target object:
  object:        (*@\ph{obj}@*)
  seed keywords: (*@\ph{keywords}@*)

List the detector keywords for the target object in this image. Return JSON now.
\end{lstlisting}
\end{rulebox}
\captionof{figure}{Selection ($\mathbb{M}_{\mathrm{sel}}$) prompt.}
\label{fig:appendix_selection_prompt}

\providecommand{\ph}[1]{{\color{teal!65!black}\itshape\{#1\}}}
\begin{rulebox}{Tool-Aware Visual Grounding $\mathbb{M}_{\mathrm{ground}}$ Prompt}
\begin{lstlisting}[basicstyle=\small,breaklines=true,breakindent=0pt,columns=fullflexible,keepspaces=true,escapeinside={(*@}{@*)}]
[SYSTEM]
You help an open-vocabulary object detector find a hard-to-detect target by
proposing better search keywords. The detector failed to confidently localize
the target with the keywords tried so far. Looking at the image, propose
alternative concrete, visually-groundable phrases (synonyms, visible parts,
superordinate or more common terms) that the detector is more likely to hit.

Return JSON only (no prose, no code fence):
{
  "keywords": ["<new phrase>", ...],   // 2-5 fresh, concrete phrases
  "confirmed": true | false,            // true if current detections already ground the target
  "reasoning": "<one short sentence>"
}

[FEW-SHOT EXEMPLARS]
<domain-specific worked examples are inserted here; omitted for brevity>

[INSTRUCTION]
Target object: (*@\ph{object}@*)
Description:   (*@\ph{description}@*)
Keywords already tried: (*@\ph{tried\_keywords}@*)
Current detector results (label, score, box): (*@\ph{detections}@*)

Propose better keywords (or confirm). Return JSON now.
\end{lstlisting}
\end{rulebox}
\captionof{figure}{Tool-Aware Visual Grounding ($\mathbb{M}_{\mathrm{ground}}$) prompt.}
\label{fig:appendix_grounding_prompt}

\end{document}